\documentclass[lettersize,journal]{IEEEtran}
\usepackage{amsmath,amsfonts}
\usepackage{algorithmic}
\usepackage{algorithm}

\usepackage{array}
\usepackage[caption=false,font=normalsize,labelfont=sf,textfont=sf]{subfig}
\usepackage{textcomp}
\usepackage{stfloats}
\usepackage{color}
\usepackage{url}
\usepackage{verbatim}
\usepackage{graphicx}
\usepackage{cite}
\usepackage{booktabs}
\usepackage{multirow}
\usepackage{amsmath}
\usepackage{amsthm}
\usepackage{amssymb}
\usepackage{tcolorbox}

\usepackage{hyperref}
\begin{document}

\title{\huge 
CaBaFL: Asynchronous Federated Learning via Hierarchical Cache and Feature Balance
}



\author{Zeke~Xia, 
        Ming~Hu,~\IEEEmembership{Member,~IEEE,}
        Dengke~Yan,
        Xiaofei~Xie,~\IEEEmembership{Member,~IEEE,}
        Tianlin~Li,~\IEEEmembership{Member,~IEEE,}
        Anran~Li,~\IEEEmembership{Member,~IEEE,}
        Junling~Zhou,~\IEEEmembership{Member,~IEEE,}
        and~Mingsong~Chen,~\IEEEmembership{Senior Member,~IEEE}
\thanks{
The authors Zeke~Xia,  Dengke~Yan, and Mingsong Chen are with the MoE Engineering Research Center of Hardware/Software Co-design Technology and Application at East China Normal University, Shanghai, 200062, China.
The authors Ming Hu and Xiaofei Xie are with Singapore Management University, Singapore.
The author Tianlin Li is with Nanyang Technological University, Singapore.
The author Anran Li is with Yale University, USA.
The author Junlong Zhou is with Nanjing University of Science and Technology, Nanjing, China.
Mingsong Chen (mschen@sei.ecnu.edu.cn) is the corresponding author.
}
}

\maketitle

\begin{abstract}
Federated Learning (FL) as a promising distributed machine learning paradigm has been widely adopted in Artificial Intelligence of Things (AIoT) applications. However, the efficiency and inference capability of FL is seriously limited due to the presence of stragglers and data imbalance across massive AIoT devices, respectively. To address the above challenges, we present a novel asynchronous FL approach named CaBaFL, which includes a hierarchical \textbf{Ca}che-based aggregation mechanism and a feature \textbf{Ba}lance-guided device selection strategy. CaBaFL maintains multiple intermediate models simultaneously for local training. The hierarchical cache-based aggregation mechanism enables each intermediate model to be trained on multiple devices to align the training time and mitigate the straggler issue. In specific, each intermediate model is stored in a low-level cache for local training and when it is trained by sufficient local devices, it will be stored in a high-level cache for aggregation. To address the problem of imbalanced data, the feature balance-guided device selection strategy in CaBaFL adopts the activation distribution as a metric, which enables each intermediate model to be trained across devices with totally balanced data distributions before aggregation. Experimental results show that compared to the state-of-the-art FL methods, CaBaFL achieves up to 9.26X training acceleration and 19.71\% accuracy improvements.
\end{abstract}


\section{Introduction}

With the prosperity of Artificial Intelligence (AI) and the Internet of Things (IoT), Artificial Intelligence of Things (AIoT) is becoming the mainstream paradigm for the design of large-scale distributed systems~\cite{xiong_iccad_xiong,han2022ads,hu_tcad_2023,tcad_zhang_2021}.
Federated Learning (FL)~\cite{fedavg,li2023towards,hu2024fedcross,hu2024fedmut,yan2023have,jia2023adaptivefl} as an important distributed machine learning paradigm has been widely used in AIoT-based applications, e.g., mobile edge computing \cite{liu2023adapterfl}, healthcare systems \cite{cvpr_quande_2021}, and autonomous driving~\cite{safavat2023asynchronous}. 
Typically, AIoT-based FL consists of a central server and a set of AIoT devices. The cloud server maintains a global model and dispatches it to multiple AIoT devices for training. Each AIoT device trains its received global model using its local data and then uploads the local model to the server. 
By aggregating all the local models, the server can achieve collaboratively global model training without leaking the raw data of any devices.

However, due to the heterogeneity of devices and data,  AIoT-based FL still encounters two main challenges. The first challenge is the straggler problem. The heterogeneous nature of AIoT devices (e.g., varying computing and wireless network communication capacities), can result in significant differences in the training time for each device~\cite{hu2020quantitative,yu2019distributed}. Aggregating all the local models, including those from devices with poor computation capabilities, can lead to longer training time. The second challenge is that the data among AIoT devices are not independent-and-identically-distributed (non-IID)~\cite{sattler_tnnls_2020}.
Such a data imbalance issue among AIoT devices can lead to the problem of ``weight divergence''~\cite{kairouz2021advances} and results in the inference accuracy degradation of the global model ~\cite{khodadadian2022federated,li2021sample,kong2021consensus,park2019wireless}. 
To address the above challenges, the existing solutions can be mainly classified into three schemes, i.e., synchronous~\cite{fedprox,icml_scaffold,zhang2023communication}, asynchronous~\cite{fedasync,fraboni2023general,shenaj2023asynchronous,hu2023gitfl}, and semi-asynchronous~\cite{safa,fedsa,zhang2023fedmds}.
In synchronous FL methods~\cite{fedavg}, the cloud server generates the global model after receiving all the local models. 
The non-IID problem could be alleviated with some well-designed training and client selection strategies \cite{fedprox,FedCluster,jin2022personalized}, while the straggler problems cannot be well addressed. Asynchronous FL methods~\cite{fedasync}  directly aggregate the uploaded local model to update the global model without waiting for other local models. By a timeout strategy, asynchronous FL methods could discard stragglers, thereby avoiding the inefficient update in the global model. However, non-IID scenarios still seriously limit the performance of existing asynchronous FL methods.
Semi-asynchronous FL methods~\cite{safa,fedsa} maintain a buffer to store uploaded local models, when the stored models reach a certain number, the server performs an aggregation operation and clears the buffer. Although semi-asynchronous FL methods can alleviate the straggler problems, they still encounter the problems of non-IID data. Due to adopting different training mechanisms, synchronous methods are often difficult to combine directly with asynchronous methods. \textit{Therefore, how to ensure performance in scenarios of data imbalance while solving the straggler problem is a serious challenge.}

To address both the straggler and data imbalance challenges, this paper presents a novel asynchronous FL approach named CaBaFL. It maintains a hierarchical cache structure to allow each intermediate model to be asynchronously trained by multiple clients before aggregation and uses a feature balance-guided client selection strategy to enable each model to eventually be trained by totally balanced data.
To address the challenge of stragglers, we use a hierarchical cache-based aggregation mechanism to achieve asynchronous training.
Specifically, we use multiple intermediate models for local training and to guarantee the number of activated devices, we enable each intermediate model participant in the aggregation after multiple times of local training.
To facilitate the asynchronous aggregation, each intermediate model is stored in the low-level cache named L2 cache. When trained with sufficient devices, a model can be stored in the high-level cache named L1 cache. While a certain number of devices trains a model, it will be updated with the global model aggregated with all the models in the L1 cache.
Based on our asynchronous mechanism, the feature balance-guided device selection strategy wisely selects a device for each intermediate model, aiming to make the total data used to train the model balanced before aggregation.
However, gaining direct access to the data distribution of each device could potentially compromise users' privacy. To address this concern, we are inspired by the observation that the middle-layer features strongly reflect the underlying data distributions  and propose a method to select devices based on their middle-layer activation patterns.
This paper has three major contributions:
\begin{itemize}
\item We propose a novel asynchronous FL framework named CaBaFL, which enables multiple intermediate models for collaborative training asynchronously using a hierarchical cache-based aggregation mechanism.



\item We present a feature balance-guided device selection strategy to wisely select devices according to the activation distribution to make each intermediate model be trained with totally balanced data, which alleviates the performance deterioration caused by data imbalance.

\item We conduct comprehensive experiments on both well-known datasets and models to show the superiority of  CaBaFL over state-of-the-art (SOTA) FL methods for both IID and non-IID scenarios. 

\end{itemize}


\section{Preliminary and Related Work}
\label{sec:related work}
\subsection{Preliminary of Federated Learning}
In general, an FL system consists of one cloud server and multiple dispersed clients. In each round of training in federated learning, the cloud server first selects a subset of devices to distribute the global model. After receiving the model, the devices conduct local training and upload the model to the cloud server. Finally, the cloud server aggregates the received models and obtains a new global model. The learning objective of FL is to minimize the loss function over the collection of training data at $N$ clients, i.e.,
\begin{equation}
\scriptsize
    \min_{w} F(w)=\sum_{k=1}^{N}\frac{\left | D_k \right | }{\left |D  \right | } F_k(w), 
\end{equation}
where $N$ is the number of clients that participate in local
training, $w$ is the global model parameters, $D_k$ is the $k^{th}$ client and  $|D_k|$ represents the training data size on $D_k$, $F_k(w)=\frac{1}{\left | D_k \right | }  {\textstyle \sum_{j\in D_k}} f_j(w)$ is the loss empirical objective over the data samples at client $k$.

\subsection{Related Work}
Asynchronous FL has a natural advantage in solving the straggler effect, where the server can aggregate without waiting for stragglers.  \cite{fedasync} develop a FedAsync algorithm, which combines a function of staleness with asynchronous update protocol. 
In FedASync, whenever a model is uploaded, the server directly aggregates it. Although FedASync can solve the straggler problem, some stragglers may become stale models, thereby reducing the accuracy of the global model. In addition, the client in FedASync will send a large number of models to the server, causing a lot of communication overhead. 
In terms of reducing data transmission, Wu et al.~\cite{safa} propose a SAFA protocol, in which asynchronous clients continuously perform local updates until the difference between the local update version and the global model version reaches tolerance.  Although SAFA considers model staleness, the server needs to wait for the asynchronous clients. Moreover, SAFA needs to maintain a bigger buffer compared to FL, which can cause more memory costs and thus lead to high complexity and low scalability.
Similarly, Ma et al. ~\cite{fedsa} set a model buffer for model aggregation to achieve semi-asynchronous FL and dynamically adjust the learning rate and local training epochs to mitigate the impact of stragglers and data heterogeneity.
However, none of the above methods can solve the problem of data heterogeneity well.

To address the problem of data heterogeneity, Zhou et al.~\cite{wkafl} propose the WKAFL protocol, which leverages the stale models of stragglers by maintaining a globally unbiased gradient and mitigates the impact of data heterogeneity through gradient clipping. 
Hu et al.~\cite{hu2023scheduling} use the semi-asynchronous FL mechanism, which maintains a buffer in the cloud server to store the local models.
The server attempts to alleviate the impact of data heterogeneity by minimizing the variance of hard labels in the buffer. However, this method requires the clients to send the hard labels to the server.
Unfortunately, in real-world scenarios, hard labels of data often contain sensitive information and cannot be obtained by the server. 
As an alternative, FedAC~\cite{fedac} employs a momentum aggregation strategy for updating the global model and incorporates fine-grained correction to adjust client gradients, effectively mitigating the challenges posed by data heterogeneity. 
FedLC~\cite{fedlc} deals with the non-IID problem by enabling local collaboration among edge devices and solves the stale model problem through dynamic learning rates. However, these methods optimize the aggregation strategy without using a wise device selection strategy, which still seriously limits their performance.
 
To the best of our knowledge, CaBaFL is the first attempt to employ collaborative training and feature balance-guided device selection strategy in asynchronous federated learning to improve both model accuracy and training stability.

\section{Motivation}\label{sec:motivation}

\subsection{Intuition of Our Asynchronous FL Mechanism}
\label{sec:3.a}
\begin{figure}[h]
\vspace{-0.1 in}
\centering
\footnotesize
\subfloat[{\scriptsize Conventional Synchronous FL}]{\includegraphics[width=0.95\linewidth]{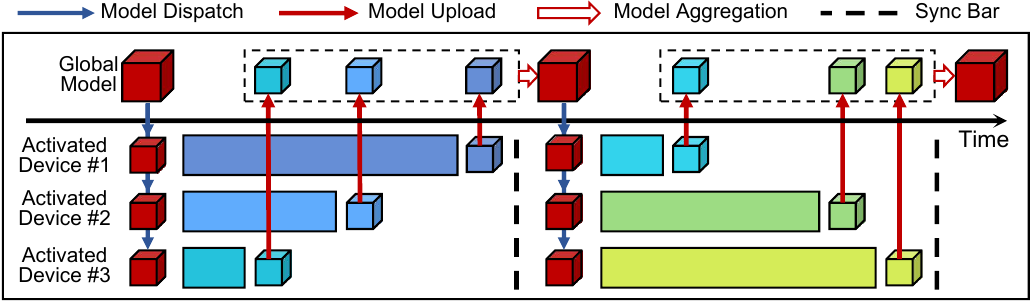}%
\label{}}
\vspace{-0.1 in}
\subfloat[{\scriptsize Intuition of our Asynchronous FL}]{\includegraphics[width=0.95\linewidth]{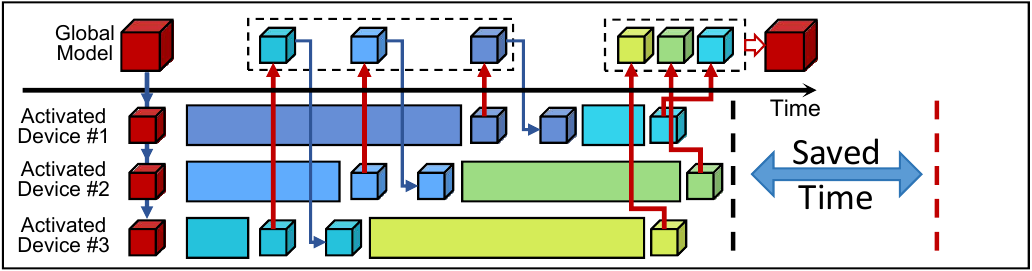}%
\label{fig:real-cifar-10-IID-ls}}
\vspace{-0.1 in}
\caption{A motivating example of our asynchronous FL method.}
\label{fig:motivation}
\vspace{-0.25 in}
\end{figure}

Figure~\ref{fig:motivation} 
presents the intuition of our asynchronous FL mechanism.
Assume that one FL training involves the local training of three models. 
Figure~\ref{fig:motivation}(a) presents a training snapshot of some conventional synchronous FL methods, where local models are aggregated after they finish training on one activated device. In this case, the FL method performs two aggregation operations without taking the straggler problem into account.
From this subfigure, we can clearly observe that some models 
become idle in between two aggregations. 
Unlike conventional FL methods, our approach always pushes models to conduct local training synchronously in a balanced way. 
Specifically, our approach considers the straggler problem and allows a model to be trained on multiple devices before an aggregation operation. 
When an activated device finishes its local training, it will immediately forward its hosting model to another device with a balanced training overhead.
As an example shown in Figure~\ref{fig:motivation}(b), a model can traverse two devices before one aggregation operation, and all three models have similar total training time spent on their two traversed devices. In this way, the straggler problem can be mitigated, since 
a compact training scheme can not only enable stragglers to be chosen for local training more often in a fair manner but also accelerate the training convergence processes. 

\subsection{Correlation between Data and Activation Distributions}

To mitigate the challenge posed by data imbalance, our objective is to carefully choose devices for each intermediate model and ensure that the total data used for training the model is balanced before aggregation. Accomplishing this task necessitates having knowledge of the data distribution associated with each device.
Due to the risk of privacy leakage, it is difficult for the server to obtain the data distribution of each device directly.
Therefore, selecting an easily obtainable metric that does not compromise privacy to guide device selection, which ensures that each intermediate model is trained by balanced data is a key challenge for our asynchronous FL approach in dealing with the non-IID problem.
We have made the following important observations that demonstrate the ability of middle-layer activation patterns to reflect the input data distribution of each device. Moreover, we have found that the activation distributions (i.e., feature distributions) 
of some model middle-layer
can provide a finer-grained representation of input data distributions than the one relying solely on input data labels. These observations motivate us to select devices based on middle-layer activation distributions.

\noindent \textbf{Observation 1.}
{\it To explore the connection between the activation distribution with the data distribution, we divide the CIFAR-10 dataset into 6 sub-datasets (i.e., $D_0-D_5$), in which the data in $D_0$ is balanced, i.e., IID, and the other five data are divided according to the Diricht distribution~\cite{measuring} $Dir(\beta)$, where a small value of $\beta$ indicates a more seriously data imbalance among sub-datasets.
We select ResNet-18 for model training and computing the activation distribution of a specific layer on the whole CIFAR-10 dataset and 6 sub-datasets respectively.} 

\begin{figure}[h]
\vspace{-0.2 in}
\centering
\footnotesize
\subfloat[{\scriptsize $\beta = 0.1$}]{\includegraphics[width=0.33\columnwidth]{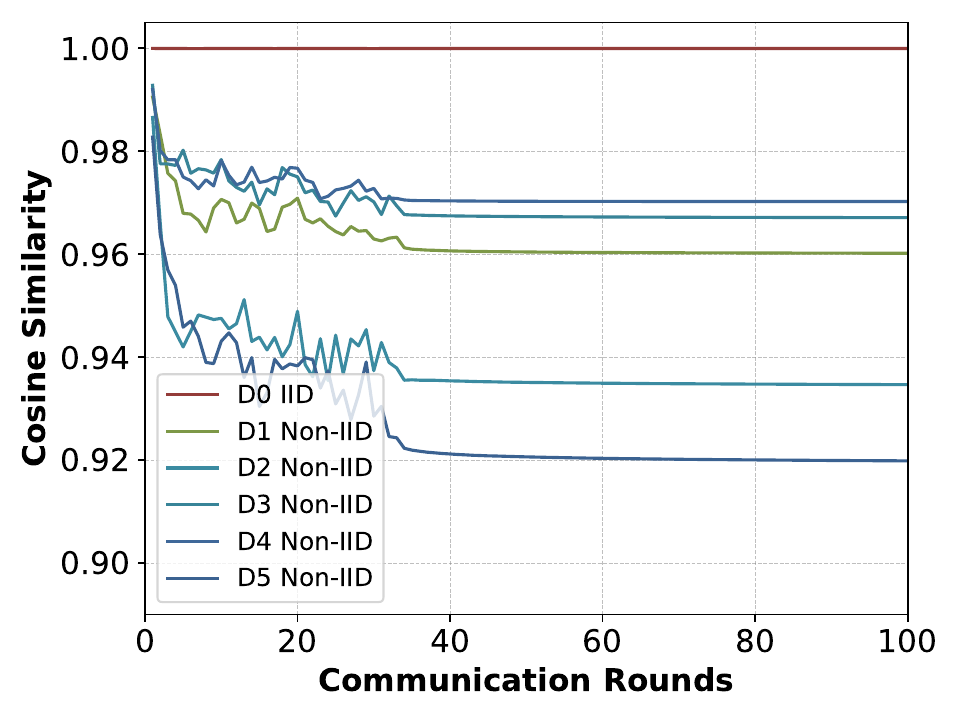}%
\label{fig:real_testbed}}
\hfil
\subfloat[{\scriptsize $\beta = 0.5$}]{\includegraphics[width=0.33\columnwidth]{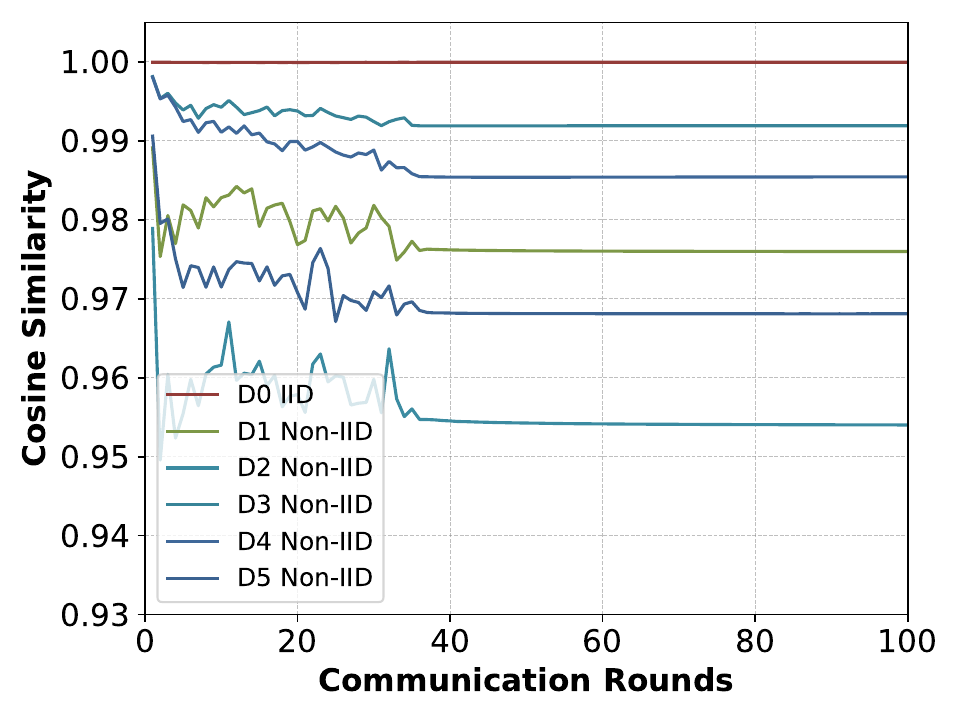}%
\label{fig:real-cifar-10-IID-ls}}
\hfil
\subfloat[{\scriptsize $\beta = 1.0$}]{\includegraphics[width=0.33\columnwidth]{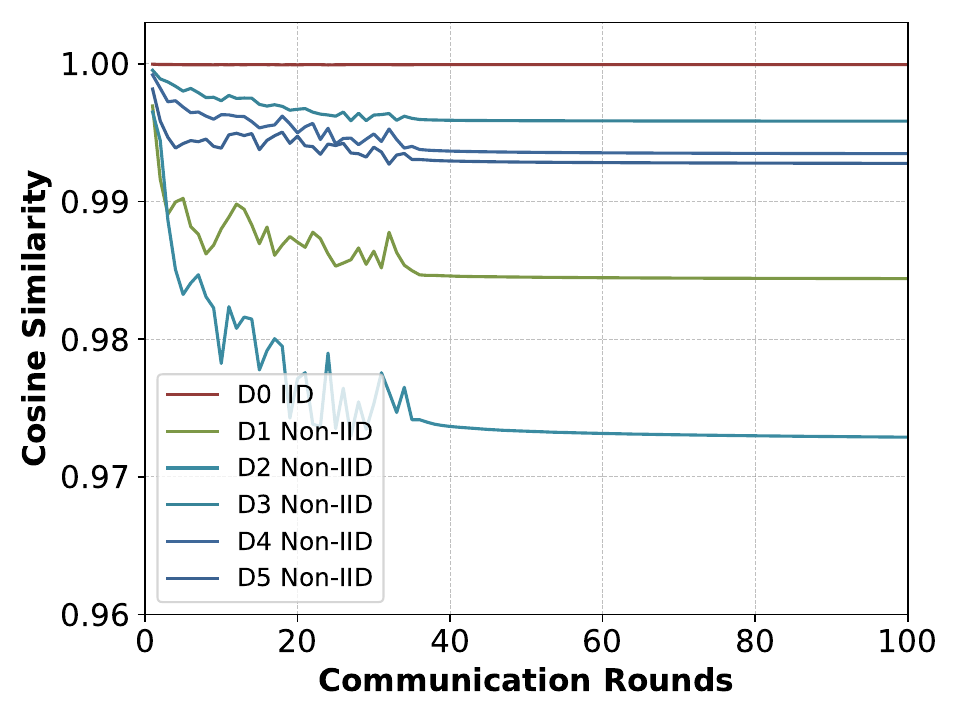}%
\label{fig:real-cifar-10-IID-lm}}
\hfil
\subfloat[{\scriptsize Combination for (a)}]{\includegraphics[width=0.33\columnwidth]{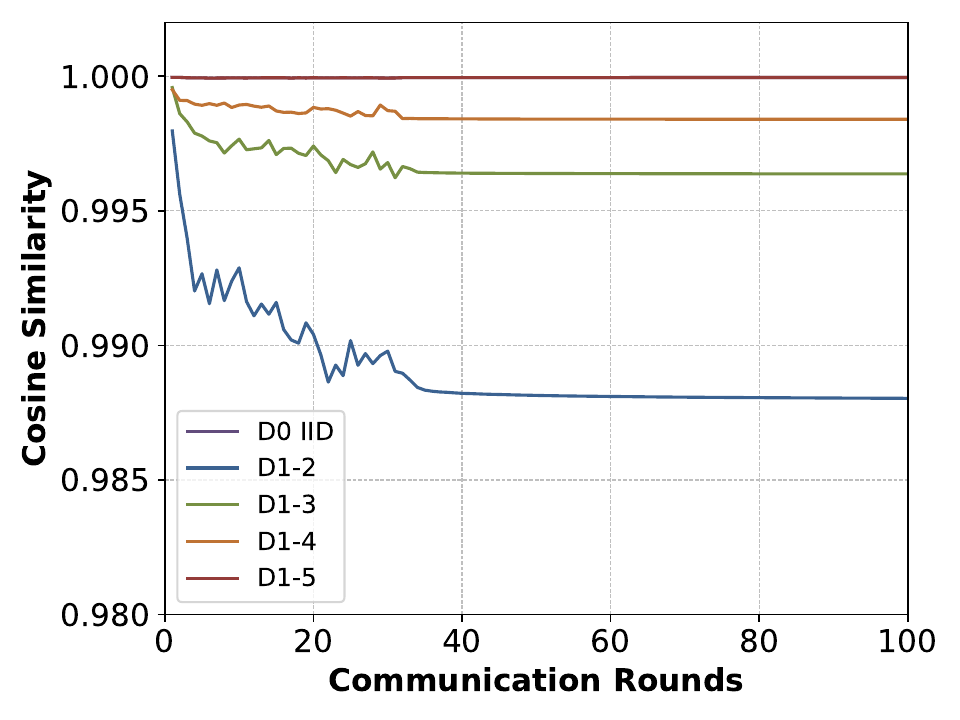}%
\label{fig:real_testbed}}
\hfil
\subfloat[{\scriptsize Combination for (b)}]{\includegraphics[width=0.33\columnwidth]{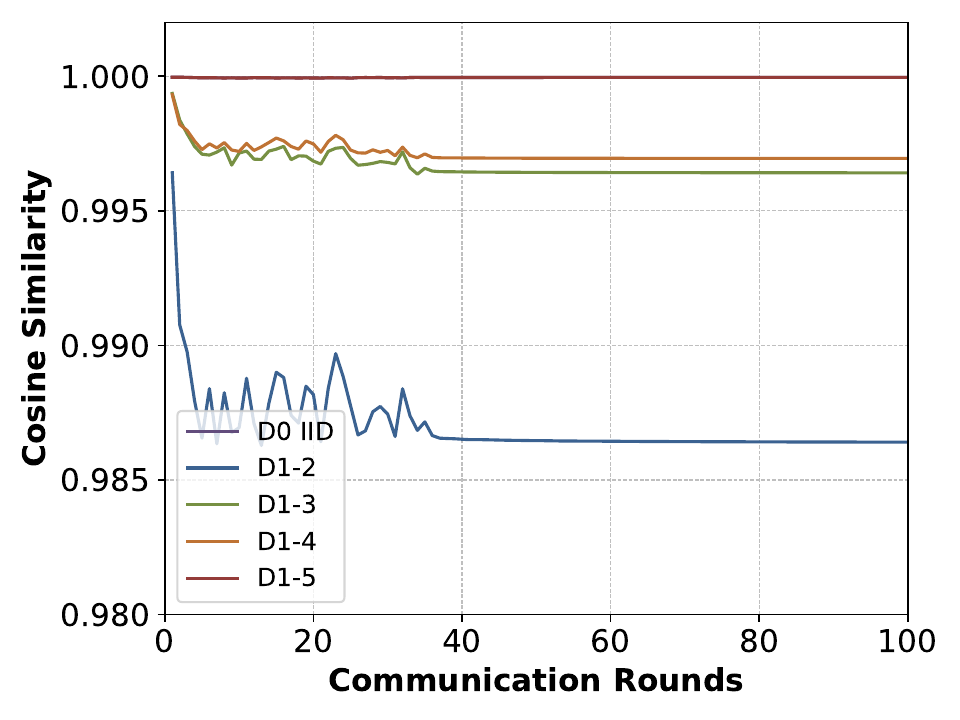}%
\label{fig:real-cifar-10-IID-ls}}
\hfil
\subfloat[{\scriptsize Combination for (c)}]{\includegraphics[width=0.33\columnwidth]{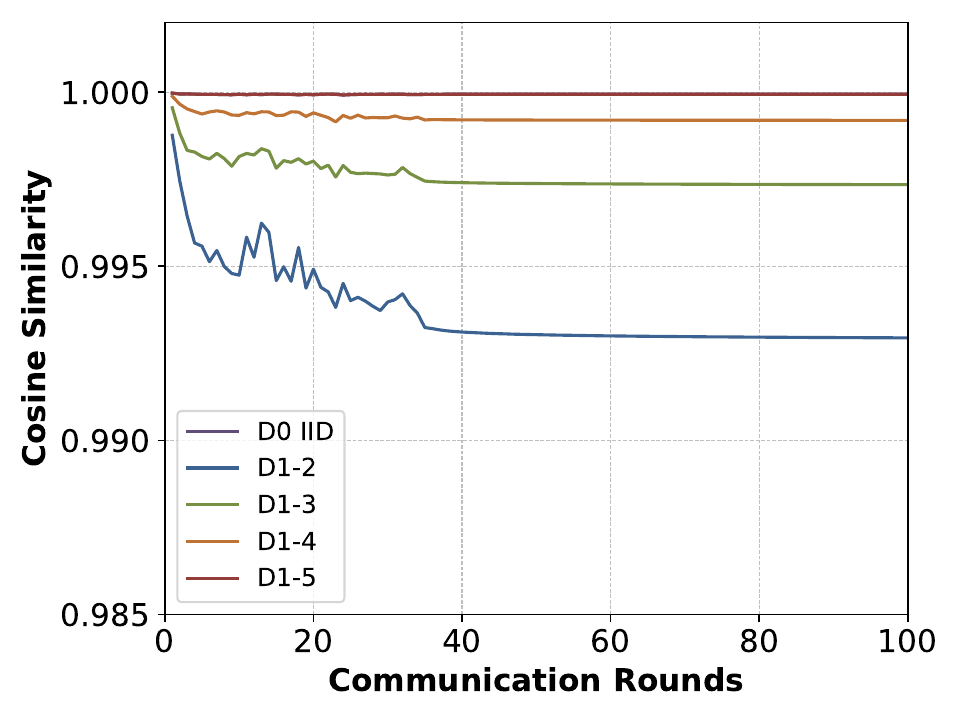}%
\label{fig:real-cifar-10-IID-lm}}
\vspace{-0.05 in}
\caption{Cosine similarity comparison of activation distributions.}
\label{fig:bigfigure}
\vspace{-0.15 in}
\end{figure}

Figures~\ref{fig:bigfigure}(a)-(c) present the cosine similarities of the activation distribution using the whole CIFAR-10 dataset with that using 6 sub-datasets, with $\beta =0.1, 0.5,$ and $1.0$, respectively.
We can observe that the $D_0$ achieves the highest cosine similarity and the sub-datasets divided with a smaller $\beta$ achieves lower cosine similarity.
Therefore, if the data distribution of a sub-dataset is more similar to that of the whole dataset, it achieves a higher cosine similarity of their activation distribution. 
Figures~\ref{fig:bigfigure}(d)-(f) present the cosine similarities of the activation distribution using the CIFAR-10 dataset with that using the combinations of 5 imbalance sub-datasets. Note that since the data in CIFAR-10 dataset and $D_0$ is balanced, data of the combination of all the 5 imbalance sub-datasets is balanced. 
From Figures~\ref{fig:bigfigure}(d)-(f), we can observe that with the data distribution of the combination dataset close to the whole dataset, the cosine similarity of their activation distribution becomes higher.
Because the activation amount is obtained by counting the number of times the neuron is activated, the activation amount of the combination dataset is equal to the sum of that of all the combined datasets.

\noindent \textbf{Observation 2.}
{\it Due to the preference difference of devices, even if they have the same data label, the features of their data may be different. For example, for the ``Dog'' category, some users prefer Husky dogs, and some users prefer Teddy dogs. Therefore, the data category distribution sometimes ignores the differences between the same category data.}

To explore whether differences between the same category data can lead to different activation distributions, we select the CIFAR-100 dataset, which contains 100 fine-grained and 20 coarse-grained classifications.
We divide the CIFAR-100 dataset into six sub-datasets (i.e., $D_0-D_5$), in which the data in $D_0$ is balanced on fine-grained classifications and the other five data are balanced on course-gained classifications but imbalanced on fine-grained classifications.
We conduct model training on the task of the coarse-grained classification and computing the activation distribution of a specific layer on the whole CIFAR-100 dataset and six sub-datasets, respectively.

\begin{figure}[h]
  \begin{center} 
		\includegraphics[width=0.3\textwidth]{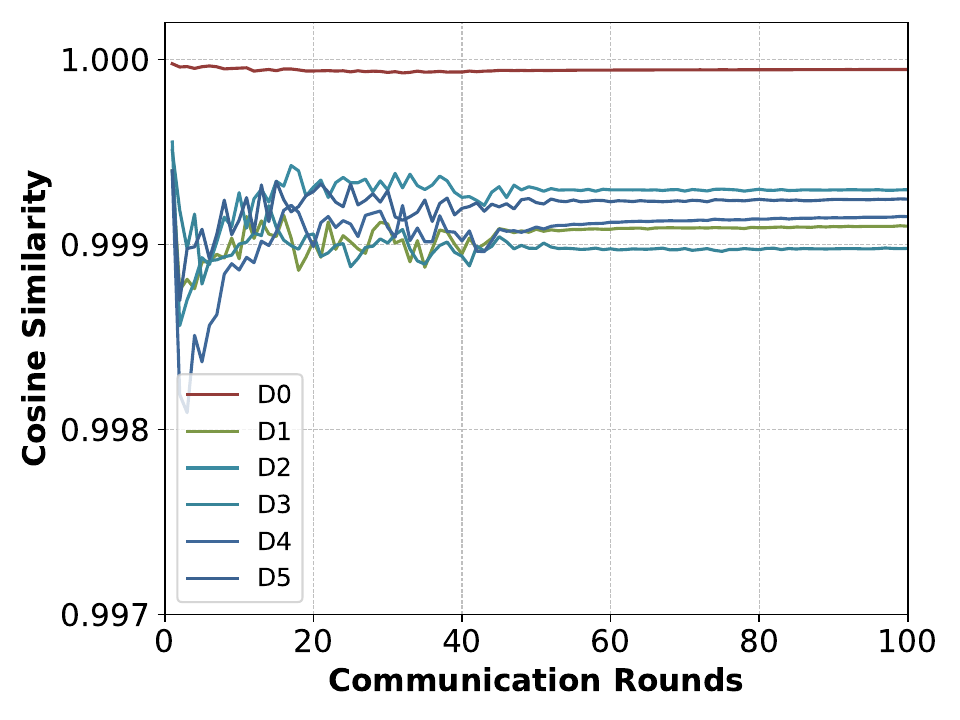}
\vspace{-0.1in}
		\caption{Cosine similarity comparison on CIFAR-100 dataset.}
		\label{fig:motivation_cifar100} 
	\end{center}
 \vspace{-0.25 in}
\end{figure}

\begin{figure*}[t] 
	\begin{center} 
		\includegraphics[width=0.8\textwidth]{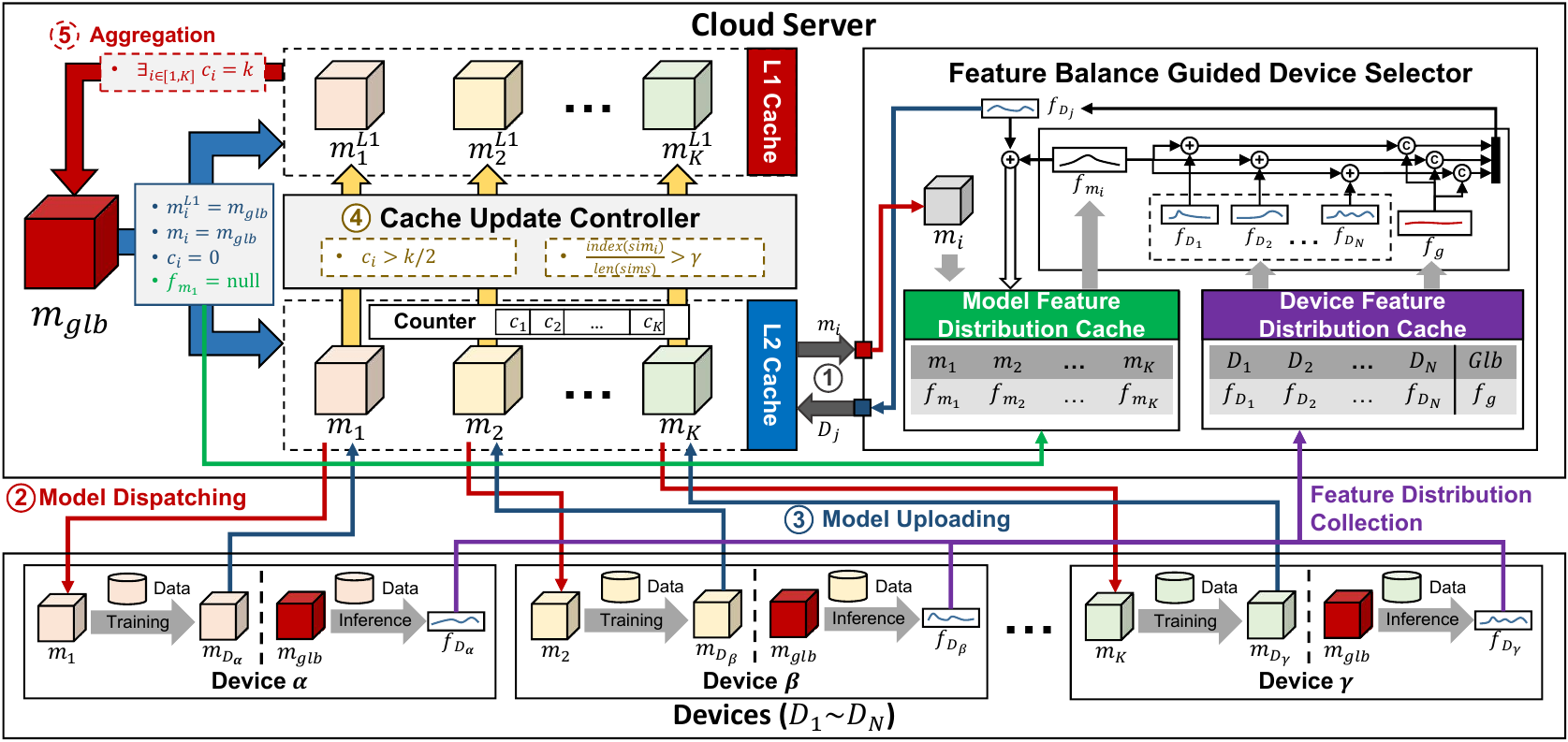}
  \vspace{-0.1in}
		\caption{Framework and workflow of CaBaFL.}
  \vspace{-0.1in}
		\label{fig:framework} 
	\end{center}
 \vspace{-0.15 in}
\end{figure*}

Figure ~\ref{fig:motivation_cifar100} presents the cosine similarities of the activation distribution using the CIFAR-100 dataset with 6 sub-datasets.
We can observe that $D_0$ can achieve the highest cosine similarity and the difference in data with the same category can result in a decrease in the similarity of activation distribution.
Therefore, compared to the data distribution, the activation distribution can express the differences between data in a more fine-grained manner.

Based on the above observations, we can find that the higher the cosine similarity 
between the activation distributions of training data and the global activation distribution, 
the more balanced the training data are. 
Therefore, selecting a device that can make the cosine similarity 
high can make the training data distribution tend to be IID. 
In our approach, we use activation distributions as a metric to guide the server to select devices to make each intermediate model trained by more balanced data.

\section{Our CaBaFL Approach}
\label{sec:method}
\subsection{Overview}
Figure~\ref{fig:framework} illustrates the framework and workflow of our CaBaFL approach, which consists of a central cloud server and multiple AIoT devices, i.e., $D_1$-$D_N$, where the cloud server includes two crucial components, i.e., a hierarchical cache-based model aggregation controller and a feature balance-guided device selector, respectively. The hierarchical cache-based model aggregation controller performs the model aggregation and updating according to the number of training times.
The feature balance-guided device selector chooses clients for local training based on the cosine similarity between the model's feature in the L2 cache and the global feature.
Note that, according to the observations in Section~\ref{sec:motivation}, we select the activation of a specific layer as the metric to calculate the feature. 
Assume that $l$ is a layer of some model $m$.
From the perspective of  $m$, 
the feature distribution of a device is the activation distribution of $l$ using raw device data as inputs of $m$.
The feature distribution of an intermediate model 
indicates the accumulative feature distributions of $l$ on all its traversed devices since the last aggregation. 
For example, assume that an intermediate model $m_k$ is continuously 
trained by devices $D_i$ and $D_j$, its feature distribution can be calculated as $f_{m_k} = f_{D_i}+f_{D_j}$, where $f_{D_i}$ and $f_{D_j}$ are the feature distribution  on $D_i$ and $D_j$ respectively. Note that, the global feature distribution is the sum of all the device feature distributions. 
Assume there are $N$ devices, the global feature can be calculated with $f_g= {\textstyle \sum_{i=1}^{N}f_{D_i}}$.

Inspired by the concept of cache in the computer architecture domain, we developed a novel two-level cache-like data structure that is hosted in the memory.
In CaBaFL, the hierarchical cache-based model aggregation controller consists of a 2-level cache data structure to store intermediate models and a cache update controller to control model updating in L1 cache.  
The feature balance-guided device selector consists of a model feature distribution cache and a device feature distribution cache, where the model feature distribution cache records the feature distribution of each intermediate model and the device feature distribution cache records the global feature distribution and feature distributions of each device.
When an intermediate model completes a local training session, the device selector updates its feature distribution by adding the feature distribution of its latest dispatched device to its original feature distribution.
In addition, CaBaFL periodically broadcasts the global model to all devices to collect their feature distributions. Upon receiving the global model, each device calculates the device feature distribution using its raw data. After the feature distribution collection, the server updates the device feature distribution cache and model feature distribution cache. Note that the feature distribution collection process operates asynchronously with the FL training process.
As shown in Figure~\ref{fig:framework}, the FL training process for each intermediate model in CaBaFL includes five steps as follows.
\begin{itemize}
\item 
\textbf{Step 1 (Device Selection}): For an intermediate model, the device selector selects a device according to the feature distribution and training time of the intermediate model and the feature distributions of candidate devices.
\item 
\textbf{Step 2 (Model Dispatching)}: The server dispatches intermediate models to  selected devices for local training.
\item 
\textbf{Step 3  (Model Uploading)}: The device uploads the model to the cloud server after local training is completed.
\item 
\textbf{Step 4  (Cache Update)}: The cache update controller stores the received model in the L2 cache and decides if it updates the model to the L1 cache according to its training times and features.
\item 
\textbf{Step 5 (Cache Aggregation)}: If an intermediate model reaches the specified training times, the cloud server aggregates all the models in the L1 cache to generate a new global model and update the intermediate model using the global model.
\end{itemize}

In our approach, steps 1-5 will be repeated until a given time threshold is reached.
Note that in the whole training process, none of the intermediate models need to wait for other intermediate models, in other words, all the intermediate models are trained asynchronously.

\vspace{-0.15 in}
\subsection{CaBaFL Implementation}
\label{sec:4.b}
\vspace{-0.05 in}
\begin{algorithm}[h]
\caption{Implementation of CaBaFL}
\label{cabafl}
\footnotesize
\raggedright \textbf{Input}:
i) $f_{g}$,  global feature distribution; ii) $f_D$,  feature distributions of devices; iii) $f_{m}$, feature distributions of models; iv) $T$, time threshold. \\
\textbf{Output}:
$m_{glb}$, the global model;\\
\textbf{CaBaFL}($f_{g}$,$f_D$,$f_m$,$T$)
\begin{algorithmic}[1]
\STATE $Init(cache, c,DS,DS^{L1}, S)$;
\STATE  $VecInit(sims)$;\\
\WHILE{time threshold $T$ is not reached}
\STATE /* Parallel for */\\
\FOR{$i\leftarrow 1,...,K$}
\STATE $m_i\leftarrow$ $ReceiveModel(i)$;
\STATE $c_i \leftarrow c_i+1$;
\STATE $cache[0][i-1]\leftarrow m_i$;
\STATE $sim_i\leftarrow cossim(f_g,f_{m_i})$;\\
\STATE $add(sims, sim_i)$;
\STATE $sort(sims)$;\\
\IF{$c_i>\frac{k}{2}$ or $\frac{index(sim_i)}{len(sims)}>\gamma$}
\STATE $cache[1][i-1]\leftarrow cache[0][i-1]$;\\
\STATE $m_i^{L1}\leftarrow cache[1][i-1]$;\\
\STATE $f_{m_i^{L1}}\leftarrow f_{m_i}$;\\
\STATE $DS^{L1}_i\leftarrow DS_i$;\\
\ENDIF
\IF{$c_i=k$}
\STATE $m_{glb} \leftarrow Aggr(m^{L1},DS^{L1})$;\\
\STATE $cache[1][i-1]\leftarrow  m_{glb}$;
\STATE $m_i\leftarrow  m_{glb}$;\\
\STATE $c_i\leftarrow 0$;\\
\STATE $f_{m_i}\leftarrow null$;\\
\ENDIF
\STATE $D_j\leftarrow DevSel(f_{g}, f_D,DS,S,c_i,f_{m_i})$;\\
\ENDFOR
\ENDWHILE
\RETURN $m_{glb}$;
\end{algorithmic}
\end{algorithm}

Algorithm \ref{cabafl} details the implementation of our CaBaFL approach. Assume that there are at most $K$ activated clients participating in local training at the same time.
Line 1 initializes the 2-level caches and other parameters we need. Note that the size of the L1 and L2 cache is $K$ intermediate models. 
Line 2 initializes the vector $sims$.
Lines 3-28 present the FL training process of models in the 2-level cache, where the “for” loop is a parallel loop. Line 6 represents that the model $m_i$ is uploaded to the server after training on the device $D_j$. Line 7 represents the training times of $m_i$ plus one. Line 8 represents the server storing the model $m_i$ in the L2 cache.
In Line 9, the server calculates the cosine similarity $sim_i$ between the model feature distribution $f_{m_i}$ and the global feature distribution $f_{g}$ as $sim_i$. After that, the server adds $sim_i$ to $sims$ and sorts $sims$ (Lines 10-11). In Lines 12-16, the server updates $m^{L1}_i$ using the eligible $m_i$. Line 13 represents that the server updates the model in the L2 cache to the corresponding L1 cache. In Lines 15-16, the server updates the feature distribution $f_{m_i^{L1}}$ and the training data size for $m_i^{L1}$ after the L1 cache is updated. Lines 18-23 present the details of the model aggregation process. In Line 18, model aggregation is triggered when $c_i$, the training times of $m_i$, achieves $k$. In Line 19, the server aggregates models in the L1 cache to obtain a new global model $m_{glb}$. In Lines 20-23, the server replaces $m_i$ and $cache[1][i-1]$ with the $m_{glb}$ and resets the training times and model feature of $m_i$. In Line 25, the function $DevSel()$ selects a device $D_j$ for $m_i$ to conduct model dispatching.

\subsubsection{Hierarchical Cache-based Asynchronous Aggregation}
\noindent \textbf{Hierarchical cache updating.}
CaBaFL maintains a 2-level cache in the cloud server to screen models in the L2 cache. When a model completes local training and is uploaded to the cloud server, the cloud server first saves the model in the L2 cache. If the model meets certain conditions,
the server updates the model to the L1 cache. The server calculate the cosine similarity $sim_i$ between the model feature distribution $f_{m_i}$ and the global feature distribution $f_{g}$ as $sim_i=cossim(f_g,f_{m_i})=\frac{f_g\cdot f_{m_i}}{\left \| f_g \right \|\left \| f_{m_i} \right \|  }.$
If $sim_i$ is low, aggregating these models hurts the performance of the global model. To perform effective model screening, the cloud server maintains a sorted list $sims$, which records the similarity between the features of all models trained and $f_g$. The server adds $sim_i$ to $sims$. $m_i^{L1}$ will be updated to $m_i$, if $\frac{index(sim_i)}{len(sims)}>\gamma$, where $index(sim_i)$ represents the index of $sim_i$ in $sims$ and $len(sims)$ represents the length of $sims$.
In addition, if the number of model training times exceeds half of the specified times, i.e., $c_i > \frac{k}{2}$, the cloud server will also update $m_i^{L1}$ to $m_i$ because the model has learned enough knowledge to participate in model aggregation.

\noindent \textbf{Model aggregation strategy (\textit{Aggr($\cdot$)})}.
In cache aggregation, our approach calculates a weight for the model in the L1 cache based on the training data size and the similarity between the model features and global features. The aggregation process is shown as follows:
\begin{equation}
\footnotesize
    Aggregation(m^{L1}, DS^{L1})= \frac{{\textstyle \sum_{i=1}^{K}}m_i^{L1}\times \frac{(DS^{L1}_i)^\alpha}{1-CS_i}}{{\textstyle \sum_{i=1}^{K}}\frac{(DS^{L1}_i)^\alpha}{1-CS_i}}, 
\end{equation}
where $m_i^{L1}$ represents the model in L1 cache, $DS^{L1}_i$ represents the size of training data of $m_i^{L1}$, $CS_i=cossim(f_{m_i^{L1}},f_g)$ represents the similarities between the feature distribution of the model in L1 cache $f_{m_i^{L1}}$ and the global feature distribution $f_g$, $K$ represents the size of the cache, and $\alpha$ is a hyperparameter that is less than 1.

To calculate the weight,  we first calculate the model weights based on the model's training data size. Due to the possibility of different numbers of training times of the model in the L1 cache, there may be significant differences in the training data size of the model. Therefore, if we directly use the training data size as the weight, it may harm the performance of the global model when data heterogeneity is high. Therefore, we use the hyperparameter $\alpha$ to mitigate this effect. After that, we calculate the weights based on the similarity $CS_i=cossim(f_g,f_{m_i^{L1}})$. The higher the $CS_i$, the more balanced the model feature distribution is, and the higher the weight is assigned to the model. Since the similarity between models is very close to 1, to achieve small differences between them, we use $1-CS_i$  to amplify these differences.

\begin{algorithm}[h]
\caption{Device Selection Procedure}
\label{selectdevice}
\footnotesize
\raggedright \textbf{Input}: i) $f_{g}$,  global feature distribution; ii) $f_D$,  feature distributions of devices;  iii) $DS$,  data size of L2 cache; iv) $S$, selected times of devices; v) $c_i$,  training times of $m_i$; vi) $f_{m_i}$,  feature distribution of $m_i$.\\
\textbf{Output}: $D_j$,  selected device.\\
\textbf{DevSel}($f_{g}$,$f_D$,$DS$,$S$,$c_i$,$f_{m_i}$)
\begin{algorithmic}[1]
\STATE $sr \leftarrow GetIdleDevice();$
\STATE $S^{idle}\leftarrow \{S_{D_i}\mid D_i\in sr\}$
\IF{$\mathrm{var}(S) > \sigma$}
\STATE $sr \leftarrow \{D_j \mid S^{idle}_{D_j}= \min(S^{idle})\}$;
\ENDIF
\IF{$c_i=0$}
\STATE $D_j\leftarrow$ $RandomlySelect(sr)$;
\RETURN $D_j$
\ENDIF
\FOR{$D_j \in sr$}
\STATE $w_1\leftarrow cossim(f_{g},f_{m_i}+f_{D_j})$;
\STATE $DS^{'}\leftarrow DS$;
\STATE $DS^{'}_i\leftarrow DS^{'}_i+\left|D_j\right|$;
\STATE $w_2 \leftarrow \mathrm{var}(DS^{'})$;
\STATE $w_{D_j}\leftarrow w_1-w_2$;
\ENDFOR
\STATE $res \leftarrow \mathop{\arg\max}\limits_{D_j} w_{D_j}$;
\STATE $S_{res}\leftarrow S_{res}+1$;
\STATE $f_{m_i}\leftarrow f_{m_i}+f_{D_{res}}$;
\STATE $DS_i \leftarrow DS_i+\left|D_{res}\right|$;
\RETURN $res$
\end{algorithmic}
\end{algorithm}
\subsubsection{Feature Balance Guided Device Selection}

Algorithm \ref{selectdevice} presents the device selection strategy of CaBaFL. When selecting devices, greedily choosing the device with the greatest similarity between model features and global features can lead to fairness issues by causing some devices to be selected repeatedly while others are rarely selected. Therefore, CaBaFL sets a hyperparameter $\sigma$. When the variance of the number of times devices are selected exceeds $\sigma$, the server selects devices from those idle devices with the fewest selections; otherwise, the selection range is all idle devices (Lines 1-4). Note that we normalize before calculating the variance.
If the model has just started a new training round, i.e., $c_i=0$, the server randomly selects a device for the model to start a new training round (Lines 6-9). 
Otherwise, for each device $D_j$, CaBaFL calculates the cosine similarity $w_1$ between the model feature distribution when $D_j$ is selected and the global feature distribution(Line 11). CaBaFL also considers balancing the total training time of the models in the L2 cache when selecting devices. Since the training time of a device is often directly proportional to the size of its data, balancing the model training data size can indirectly balance training time between models in the L2 cache.
More balanced training times between models can alleviate the stale model problem caused by the stragglers and allow for faster training.
Therefore, CaBaFL calculates the variance of the training data size of the model in the L2 cache when $D_j$ is selected (Lines 12-14), where $|D_j|$ represents the training data size on $D_j$. $w_1$ is then subtracted by the variance to obtain the weight of the device (Line 15), and the device with the highest weight is selected for model dispatching (Line 17). Finally, Lines 18-20 update $S_{res}$, $f_{m_i}$ and $DS_i$.

\begin{table*}[h]
\centering
\vspace{-0.05 in}
\caption{Comparison of test accuracy for both IID and non-IID scenarios.}
\label{table:acc_compare}
\vspace{-0.1 in}
\resizebox{\textwidth}{!}{%
\begin{tabular}{cccccccccc}
\toprule
\multirow{2}{*}{Dataset}    & \multirow{2}{*}{Model}     & \multirow{2}{*}{\begin{tabular}[c]{@{}c@{}}Heterogeneity\\ Settings\end{tabular}} & \multicolumn{7}{c}{Test Accuracy (\%)}                                                                               \\ \cline{4-10} 
                            &                            &                                                                                   & FedAvg~\cite{fedavg}         & FedProx~\cite{fedprox}        & FedASync~\cite{fedasync}       & FedSA~\cite{fedsa}          & SAFA~\cite{safa}           & WKAFL~\cite{wkafl}          & CaBaFL         \\ \hline
\multirow{12}{*}{CIFAR-10}  & \multirow{4}{*}{ResNet-18} & $\beta=0.1$                                                                             & $45.65\pm3.52$ & $45.57\pm1.42$ & $47.34\pm0.72$ & $47.26\pm1.61$ & $42.20\pm2.63$ & $40.86\pm5.98$ & \boldmath{$55.46\pm0.67$} \\
                            &                            & $\beta=0.5$                                                                             & $60.34\pm0.37$ & $58.66\pm0.27$ & $60.70\pm0.28$ & $60.59\pm0.28$ & $57.69\pm0.71$ & $67.84\pm0.78$ & \boldmath{$69.31\pm0.18$} \\
                            &                            & $\beta=1.0$                                                                             & $65.79\pm0.34$ & $63.55\pm0.11$ & $66.61\pm0.22$ & $65.11\pm0.27$ & $61.92\pm0.44$ & $71.19\pm0.49$ & \boldmath{$72.84\pm0.36$} \\
                            &                            & $IID$                                                                               & $64.89\pm0.17$ & $63.78\pm0.14$ & $64.76\pm0.10$ & $64.98\pm0.12$ & $64.28\pm0.16$ & $73.50\pm0.18$ & \boldmath{$74.83\pm0.07$} \\ \cline{2-10} 
                            & \multirow{4}{*}{CNN}       & $\beta=0.1$                                                                             & $46.58\pm1.29$ & $46.59\pm1.06$ & $48.90\pm0.68$ & $48.58\pm1.92$ & $42.66\pm2.23$ & $41.03\pm3.91$ & \boldmath{$52.92\pm0.62$} \\
                            &                            & $\beta=0.5$                                                                             & $54.77\pm0.60$ & $54.33\pm0.22$ & $55.97\pm0.66$ & $55.44\pm0.37$ & $52.63\pm1.30$ & $50.25\pm3.15$ & \boldmath{$57.89\pm0.59$} \\
                            &                            & $\beta=1.0$                                                                             & $55.88\pm0.53$ & $55.72\pm0.28$ & $56.00\pm0.31$ & $56.45\pm0.28$ & $54.42\pm1.02$ & $53.88\pm1.78$ & \boldmath{$58.58\pm0.72$} \\
                            &                            & $IID$                                                                              & $57.87\pm0.19$ & $57.76\pm0.21$ & $57.82\pm0.09$ & $57.94\pm0.14$ & $55.88\pm0.17$ & $58.15\pm0.38$ & \boldmath{$61.75\pm0.30$} \\ \cline{2-10} 
                            & \multirow{4}{*}{VGG-16}    & $\beta=0.1$                                                                            & $62.67\pm5.15$ & $63.31\pm3.28$ & $66.26\pm1.24$ & $65.26\pm2.13$ & $56.37\pm3.95$ & $55.09\pm3.92$ & \boldmath{$70.68\pm1.49$} \\
                            &                            & $\beta=0.5$                                                                             & $77.94\pm0.46$ & $77.15\pm0.09$ & $78.26\pm0.20$ & $78.02\pm0.26$ & $75.40\pm0.87$ & $78.78\pm0.20$ & \boldmath{$82.36\pm0.25$} \\
                            &                            & $\beta=1.0$                                                                             & $79.45\pm0.41$ & $78.58\pm0.14$ & $79.69\pm0.17$ & $78.64\pm0.33$ & $77.46\pm0.75$ & $80.19\pm1.02$ & \boldmath{$83.81\pm0.27$} \\
                            &                            & $IID$                                                                               & $80.35\pm0.05$ & $79.22\pm0.08$ & $80.30\pm0.06$ & $80.42\pm0.04$ & $79.25\pm0.14$ & $82.89\pm0.24$ & \boldmath{$85.12\pm0.05$} \\ \hline
\multirow{12}{*}{CIFAR-100} & \multirow{4}{*}{ResNet-18} & $\beta=0.1$                                                                             & $33.64\pm0.52$ & $33.24\pm0.38$ & $35.05\pm0.41$ & $33.61\pm0.49$ & $31.01\pm1.04$ & $27.65\pm2.22$ & \boldmath{$37.77\pm0.35$} \\
                            &                            & $\beta=0.5$                                                                             & $41.31\pm0.19$ & $40.08\pm0.18$ & $42.75\pm0.37$ & $41.96\pm0.20$ & $39.44\pm0.44$ & $44.78\pm1.48$ & \boldmath{$47.30\pm0.28$} \\
                            &                            & $\beta=1.0$                                                                             & $43.33\pm0.19$ & $41.76\pm0.11$ & $44.85\pm0.23$ & $42.75\pm0.23$ & $40.92\pm0.23$ & $47.41\pm0.76$ & \boldmath{$48.10\pm0.44$} \\
                            &                            & $IID$                                                                               & $42.76\pm0.09$ & $42.63\pm0.15$ & $42.20\pm0.18$ & $43.18\pm0.15$ & $42.07\pm0.16$ & $49.10\pm0.20$ & \boldmath{$49.64\pm0.06$} \\ \cline{2-10} 
                            & \multirow{4}{*}{CNN}       & $\beta=0.1$                                                                             & $29.44\pm0.65$ & $30.64\pm0.83$ & $31.81\pm0.57$ & $30.52\pm0.49$ & $29.44\pm1.44$ & $19.18\pm2.59$ & \boldmath{$33.22\pm0.48$} \\
                            &                            & $\beta=0.5$                                                                             & $34.08\pm0.64$ & $34.73\pm0.45$ & $32.99\pm0.30$ & $34.70\pm0.49$ & $33.98\pm0.95$ & $28.88\pm2.25$ & \boldmath{$36.44\pm0.39$} \\
                            &                            & $\beta=1.0$                                                                             & $32.43\pm0.48$ & $32.74\pm0.38$ & $32.84\pm0.27$ & $34.27\pm0.30$ & $33.55\pm0.35$ & $31.35\pm1.35$ & \boldmath{$37.31\pm0.49$} \\
                            &                            & $IID$                                                                               & $33.04\pm0.24$ & $33.55\pm0.20$ & $32.60\pm0.22$ & $33.23\pm0.15$ & $31.73\pm0.37$ & $32.99\pm0.68$ & \boldmath{$36.88\pm0.33$} \\ \cline{2-10} 
                            & \multirow{4}{*}{VGG-16}    & $\beta=0.1$                                                                             & $47.30\pm1.27$ & $47.29\pm0.45$ & $49.69\pm0.68$ & $46.91\pm1.41$ & $42.34\pm1.66$ & $30.82\pm3.78$ & \boldmath{$50.53\pm0.49$} \\
                            &                            & $\beta=0.5$                                                                             & $54.83\pm0.59$ & $53.80\pm0.59$ & $56.39\pm0.47$ & $54.55\pm0.40$ & $50.61\pm0.66$ & $53.08\pm2.40$ & \boldmath{$57.37\pm0.37$} \\
                            &                            & $\beta=1.0$                                                                             & $56.49\pm0.21$ & $54.50\pm0.35$ & $57.30\pm0.35$ & $55.24\pm0.31$ & $53.65\pm0.40$ & $56.71\pm1.07$ & \boldmath{$59.53\pm0.17$} \\
                            &                            & $IID$                                                                               & $57.56\pm0.05$ & $56.30\pm0.23$ & $57.91\pm0.13$ & $56.39\pm0.14$ & $55.33\pm0.24$ & $61.16\pm0.38$ & \boldmath{$64.96\pm0.04$} \\ \hline
\multirow{3}{*}{FEMNIST}    & ResNet-18                  & -                                                               & $82.25\pm0.43$ & $81.57\pm0.30$ & $82.87\pm0.35$ & $82.07\pm0.43$ & $76.85\pm0.35$ & $74.68\pm0.60$ & \boldmath{$84.08\pm0.17$} \\ \cline{2-10} 
                            & CNN                        &   -                                                                                & $81.29\pm0.38$ & $81.57\pm0.29$ & $82.68\pm0.23$ & $81.99\pm0.46$ & $76.82\pm0.38$ & $73.81\pm0.58$ & \boldmath{$84.00\pm0.10$} \\ \cline{2-10} 
                            & VGG-16                     &  -                                                                                 & $82.34\pm0.38$ & $81.38\pm0.21$ & $82.73\pm0.26$ & $82.00\pm0.47$ & $76.87\pm0.39$ & $74.62\pm0.47$ & \boldmath{$84.30\pm0.13$} \\ \hline
\end{tabular}%
}
\vspace{-0.2 in}
\end{table*}

\section{Performance Evaluation}
\label{sec:exp}
\subsection{Experimental Setup}
\vspace{-0.05 in}
\label{sec:5.a}
To demonstrate the effectiveness of our approach, we implemented CaBaFL using the PyTorch framework~\cite{pytorch}. All experimental results were obtained from a Ubuntu workstation equipped with an Intel i9 CPU, 64GB of memory, and an NVIDIA RTX 4090 GPU.

\noindent \textbf{Settings of Baselines.} We compared our approach with six baselines, including
the classical FedAvg~\cite{fedavg} and five SOTA FL methods (i.e., FedProx~\cite{fedprox}, FedAsync~\cite{fedasync}, SAFA~\cite{safa}, FedSA~\cite{fedsa}, and WKAFL~\cite{wkafl}), which aim to solve similar problem. 
Note that FedAvg and FedProx are synchronous FL methods, FedAsync is an asynchronous FL method, and the other three are semi-asynchronous FL methods.
For SAFA, FedSA, and WKAFL, we set the model buffer size to $K/2$. 
To ensure a fair comparison, we used an SGD optimizer with a learning rate of 0.01 and a momentum of 0.5 for all baselines and CaBaFL. Each client was trained locally for five epochs with a batch size of 50. Note that all these experimental settings are widely used in the evaluation of baselines.

\noindent \textbf{Settings of Datasets and Models.}
Our experiments were conducted on three well-known datasets, i.e., CIFAR-10, CIFAR-100~\cite{cifar10}, and FEMNIST~\cite{femnist}, which are widely used in evaluating the performance of the above baselines.
To create the heterogeneity of device data for CIFAR-10 and CIFAR-100, we employed the Dirichlet distribution $Dir(\beta)$ \cite{measuring}, where smaller values of $\beta$ indicate greater data heterogeneity. As FEMNIST already possesses a non-IID distribution, we did not require the use of the Dirichlet distribution. Moreover, to show the pervasiveness of our approach, we conducted experiments on three well-known
networks, i.e., CNN 
, ResNet-18 
, and VGG-16 
, respectively, which have different structures and depths.

\noindent \textbf{Settings of System Heterogeneity.}
To simulate system heterogeneity, for the experiments on datasets CIFAR-10 and CIFAR-100, we assumed each of them involved 100 AIoT devices with varying computing power. However, for dataset FEMNIST, there are a total of 180 devices.
 Firstly, we simulate different computing power based on the processing speed of true devices. Since the NVIDIA Jetson AGX Xavier device can be powered by different performance modes that can provide different computing power, we measured the processing speed under different performance modes and used it as the basis for simulating the computing power of our devices. To generate the simulated computing power, we used a Gaussian distribution with the mean and variance obtained from actual time consumption data of training models on Jetson AGX Xavier. Specifically, we assume that the training performance  (i.e., the training time of one data sample) of a device follows the Gaussian distribution $N(0.03, 0.01)$, as measured in seconds. Secondly, we simulated a fixed network bandwidth for each device to calculate the communication time required with the cloud server. In addition, we assume only 10\% of devices can participate in training at the same time.

\subsection{Performance Comparison}
\subsubsection{Comparison of Accuracy}
\begin{figure}[h]
\vspace{-0.2 in}
\centering
\footnotesize
\subfloat[{\small $\beta = 0.1$}]{\includegraphics[width=0.4\columnwidth]{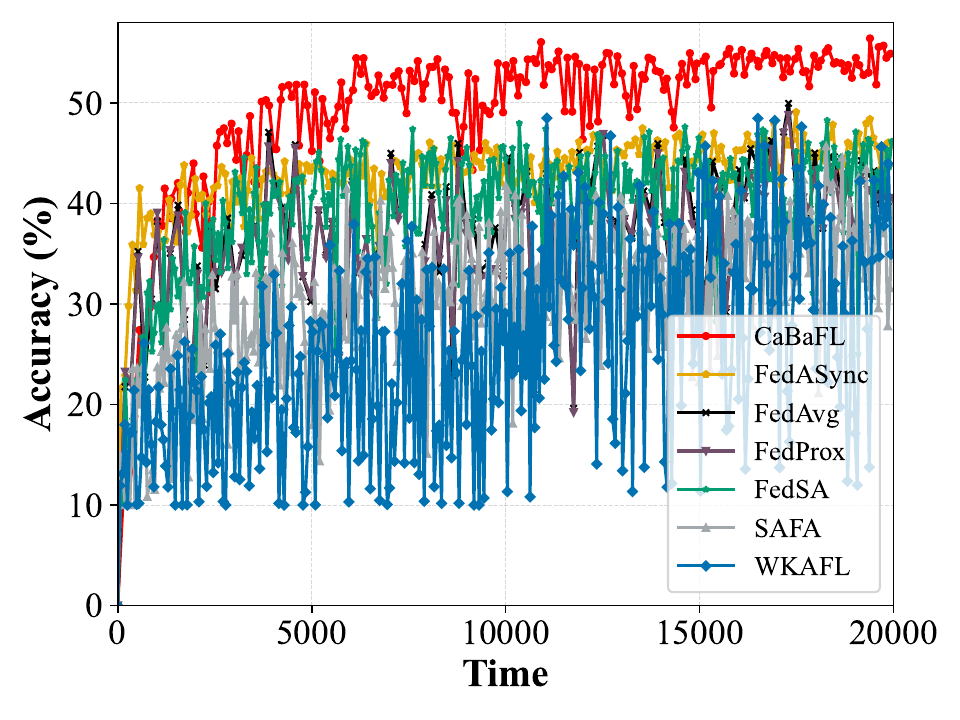}%
\label{}}
\hfil
\subfloat[{\small $\beta = 0.5$}]{\includegraphics[width=0.4\columnwidth]{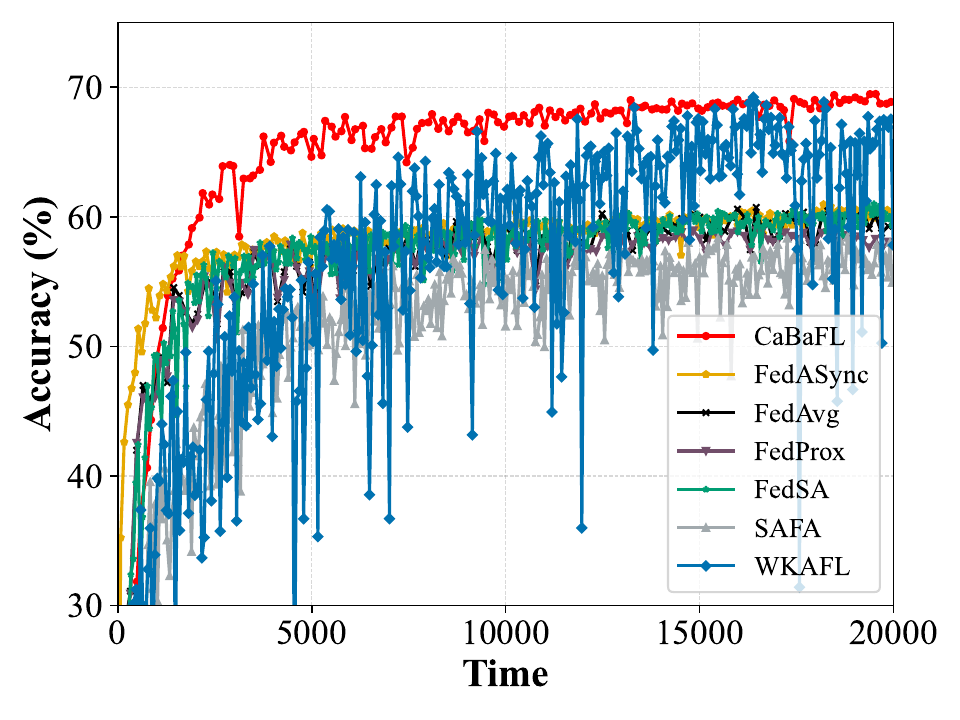}%
\label{fig:real-cifar-10-IID-ls}}
\hfil
\subfloat[{\small $\beta = 1.0$}]{\includegraphics[width=0.4\columnwidth]{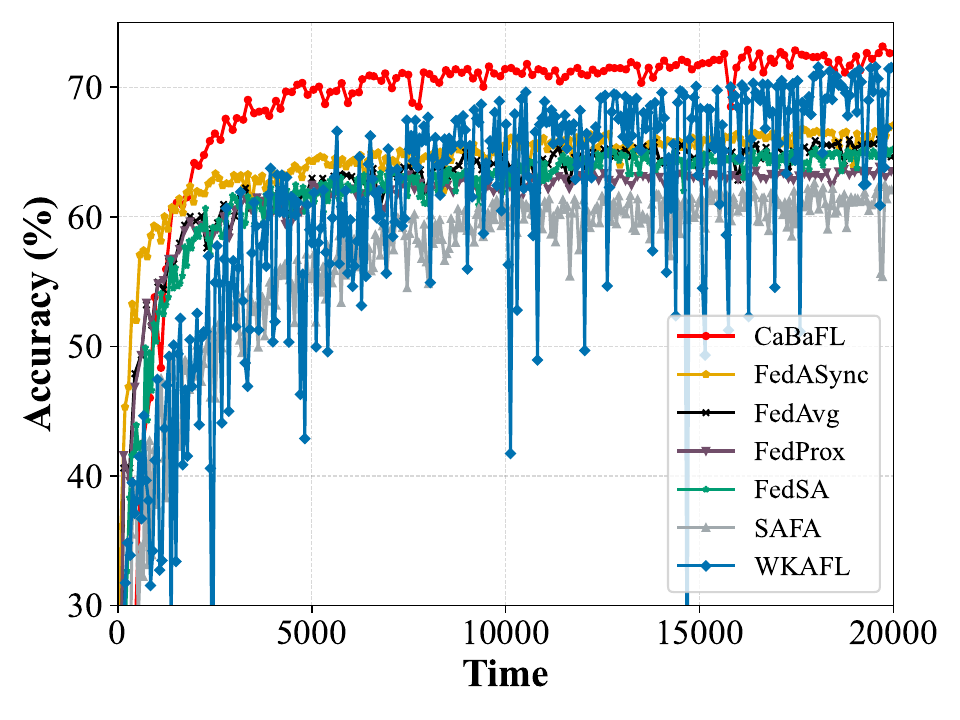}%
\label{fig:real-cifar-10-IID-lm}}
\hfil
\subfloat[{\small IID}]{\includegraphics[width=0.4\columnwidth]{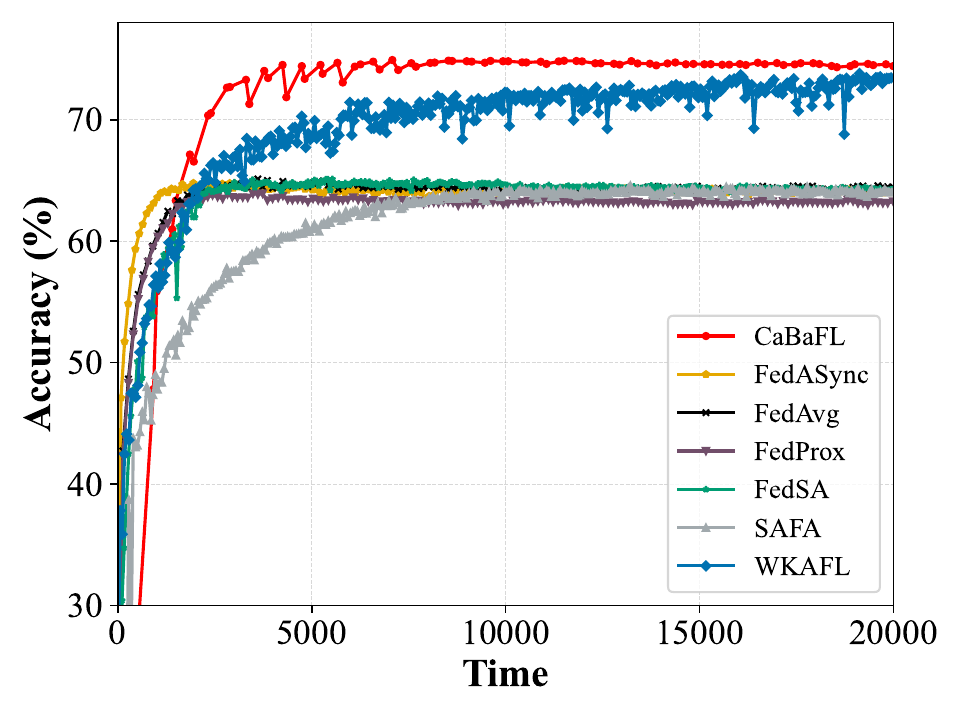}%
\label{}}
\vspace{-0.1 in}
\caption{Learning curves of training CIFAR-10 with ResNet-18.}
\label{fig:acc_compare}
\vspace{-0.15 in}
\end{figure}

Table \ref{table:acc_compare} compares the test accuracy between CaBaFL and all the baselines on three datasets with different non-IID and IID settings using ResNet-18, CNN, and VGG-16 models. From Table \ref{table:acc_compare}, it can be observed that CaBaFL can achieve the highest test accuracy in all the scenarios regardless of model type, dataset type, and data heterogeneity. For example, CaBaFL improves test accuracy by 19.71\% over WKAFL in CIFAR-100 dataset with VGG-16 model when $\beta=0.1$.  Figure \ref{fig:acc_compare} shows the learning curves of CaBaFL and all baseline methods on CIFAR-10 and ResNet-18. As an example of dataset CIFAR-10, when $\beta$ = 0.1 and the target accuracy is 45\%, CaBaFL outperforms SAFA by 9.26X in terms of training time. Moreover, We can observe that CaBaFL achieves the highest accuracy and exhibits good stability in its learning curve. 
Furthermore, we can find that our method can perform better than the baselines even in the IID scenario. This is mainly because in CaBaFL, a model can be trained on multiple devices before being aggregated, i.e., it can perform more times of stochastic gradient descent. 

\begin{table}[h]\setlength\tabcolsep{3pt}
\centering
\vspace{-0.1 in}
\caption{Comparison of communication overhead.}
\label{comm}
\vspace{-0.1 in}
\resizebox{\linewidth}{!}{%
\begin{tabular}{ccccccccc}
\hline
\multirow{2}{*}{\begin{tabular}[c]{@{}c@{}}Heter.\\ Settings\end{tabular}} & \multirow{2}{*}{Acc. (\%)} & \multicolumn{7}{c}{Communication Overhead} \\ \cline{3-9} 
                         &    & FedAvg & FedProx & FedASync     & FedSA & SAFA  & WKAFL & CaBaFL        \\ \hline
\multirow{2}{*}{$\beta=0.1$} & 40 & 1620   & 1620    & \textbf{600} & 1510  & 4700  & 7930  & 1360          \\ \cline{2-9} 
                         & 45 & 2300   & 2300    & 4220         & 2380  & 14640 & 7940  & \textbf{2120} \\ \hline
\multirow{2}{*}{$\beta=0.5$} & 58 & 2760   & 2760    & 3160         & 3160  & 9550  & 2680  & \textbf{1811} \\ \cline{2-9} 
                         & 60 & 7860   & NA       & 11060        & 11040 & NA     & 3380  & \textbf{2259} \\ \hline
\multirow{2}{*}{$\beta=1.0$} & 63 & 2920   & 5200    & 2580         & 4560  & NA     & 2960  & \textbf{2085} \\ \cline{2-9} 
                         & 65 & 7480   & NA       & 5760         & 10270 & NA     & 3270  & \textbf{2279} \\ \hline
\multirow{2}{*}{$IID$}   & 63 & 1200   & 1220    & \textbf{940} & 1770  & 5800  & 1580  & 1593          \\ \cline{2-9} 
                         & 70 & NA      & NA       & NA            & NA     & NA     & 4010  & \textbf{2441} \\ \hline
\end{tabular}
}
\vspace{-0.2 in}
\end{table}

\subsubsection{Comparison of Communication Overhead}
To evaluate the communication overhead caused by CaBaFL, we 
conducted experiments on the CIFAR-10 dataset using the ResNet-18 model. Table \ref{comm} compares the communication overheads between CaBaFL and the baselines to achieve specified inference accuracy for the global model. 
We can find CaBaFL leads to the least communication overhead
within six out of the eight cases.
For example, when the target accuracy is 65\% and $\beta=1.0$, 
the communication overhead of FedASync is 5760, while our approach is 2279. Note that both FedProx and SAFA cannot achieve the target in this case.

\subsubsection{Comparison of Stability and Fairness}
We conducted experiments to quantify the stability and fairness of models 
using the CIFAR-10 dataset and ResNet-18 model with $\beta=0.5$.
For stability analysis,  we used the standard deviation of the Moving Average (MA) to quantify model stability. We calculated the performance of our method and all baselines on this metric, and the results are shown in Table \ref{tab:stable}. We can find that CaBaFL performs best. 
Meanwhile, our approach considers the fairness of device selection. 
By introducing the hyperparameter $\sigma$ (see Line 3 of Algorithm 2), our device selection strategy can ensure that all devices are selected with a similar number of times. We
normalized the number of times a device is selected and used
the variance of the normalized values of devices to quantify such fairness. We compared our device selection strategy with the classic random device selection strategy. 
We found that the fairness of using our device selection strategy is $8.7\times 10^{-7}$, while the fairness of using the random strategy is $1\times10^{-6}$, indicating the fairness of CaBaFL is similar to the one of FedAvg.   

\begin{table}[h]
\centering
\vspace{-0.15 in}
\caption{Stability comparison using the Standard Deviation of MA.}
\vspace{-0.1 in}
\label{tab:stable}
\resizebox{\linewidth}{!}{%
\begin{tabular}{|ccccccc|}
\hline
\multicolumn{1}{c}{FedAvg} & \multicolumn{1}{c}{FedProx} & \multicolumn{1}{c}{FedASync} & \multicolumn{1}{c}{FedSA} & \multicolumn{1}{c}{SAFA} & \multicolumn{1}{c}{WKAFL} & \multicolumn{1}{c}{CaBaFL} \\ \hline
\multicolumn{1}{c}{0.72} & \multicolumn{1}{c}{0.41} & \multicolumn{1}{c}{0.38} & \multicolumn{1}{c}{0.56} & \multicolumn{1}{c}{1.47} & \multicolumn{1}{c}{4.13} & \multicolumn{1}{c}{\textbf{0.37}}\\ \hline
\end{tabular}%
}
\vspace{-0.25 in}
\end{table}
\subsection{Impacts of Different Configuration}
\label{sec:5.c}
\subsubsection{Impacts of Different Features}
\begin{figure}[h]
\vspace{-0.2 in}
\centering
\footnotesize
\subfloat[{\small $\beta = 0.5$}]{\includegraphics[width=0.4\columnwidth]{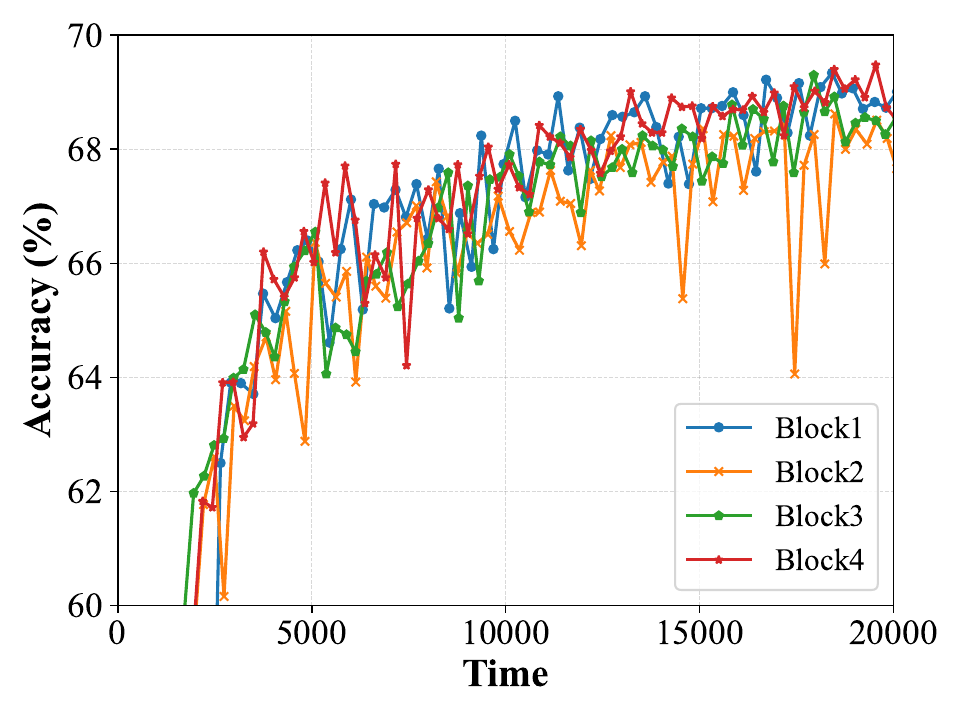}%
\label{}}
\hfil
\subfloat[{\small IID}]{\includegraphics[width=0.4\columnwidth]{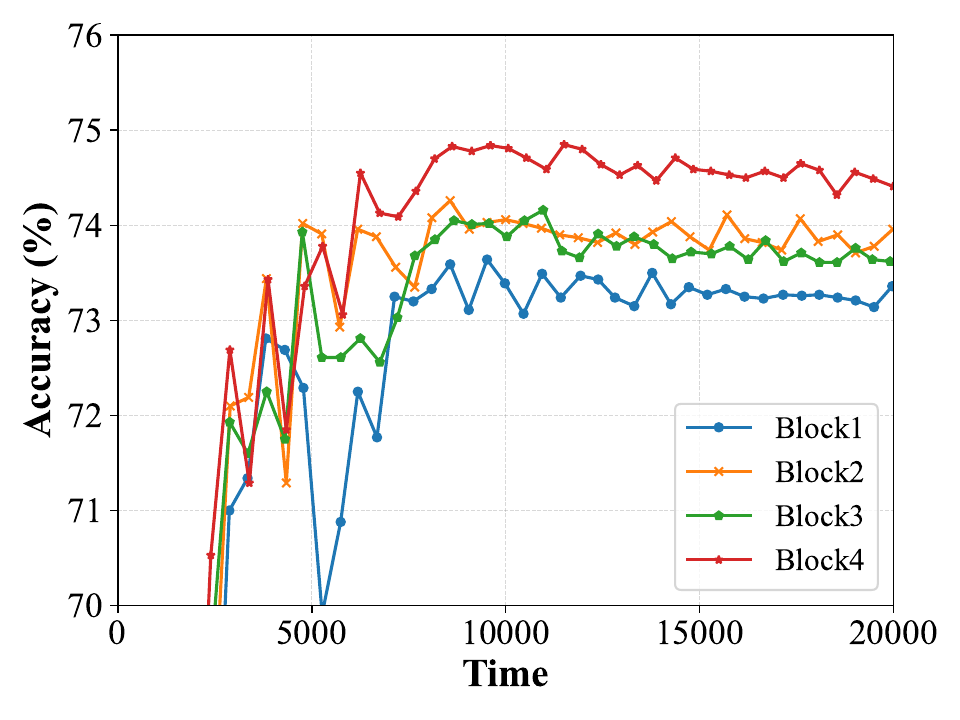}%
\label{fig:real-cifar-10-IID-ls}}
\vspace{-0.05 in}
\caption{Impact of different features on test accuracy.}
\label{fig:feature_selection}
\vspace{-0.15 in}
\end{figure}

We investigated the impact of selecting features from different model layers on accuracy. Since ResNet-18 has 4 blocks, we performed pooling operations on the feature outputs of each block separately and used them as features when selecting the device. The feature dimensions of Blocks 1 to 4 are 64, 128, 256, and 512, respectively. Figure \ref{fig:feature_selection} shows all experiment results conducted on CIFAR-10 with IID distribution and Dirichlet distribution where $\beta=0.5$. We can observe that selecting the features from Block 4 performs the best. We can conclude that because of the highest dimensionality of Block 4, the features from Block 4 can represent data distribution in a more fine-grained manner, thus achieving the highest accuracy.

\subsubsection{Impacts of Different Number of Simultaneously Training Devices}
\begin{figure}[h]
\vspace{-0.15 in}
\centering
\footnotesize
\subfloat[{\small $5\%$}]{\includegraphics[width=0.4\columnwidth]{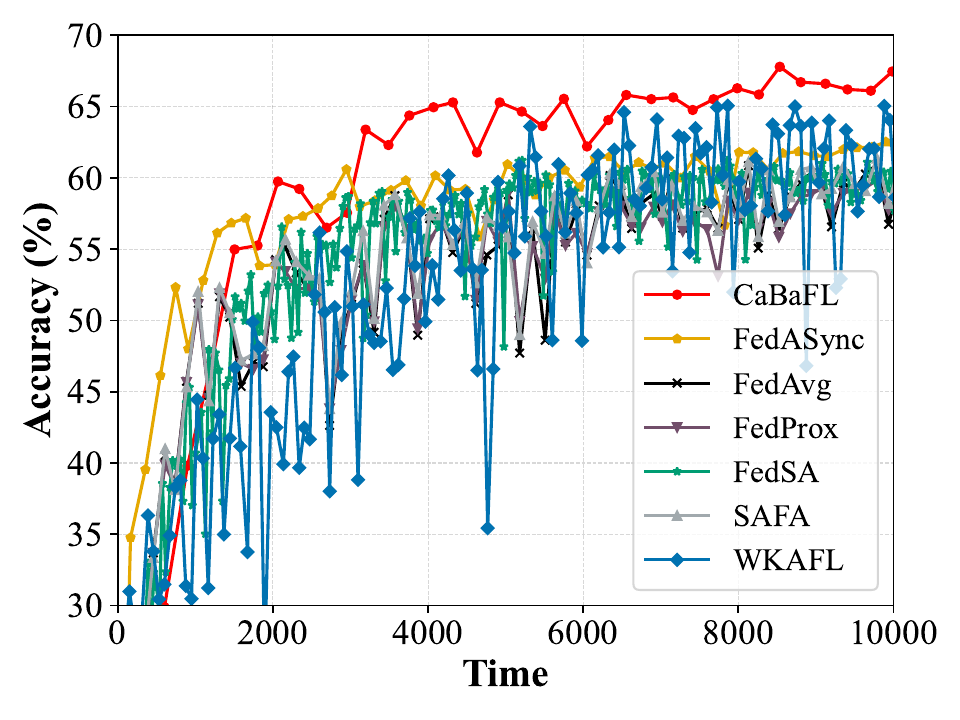}%
\label{fig:real_testbed}}
\hfil
\subfloat[{\small $10\%$}]{\includegraphics[width=0.4\columnwidth]{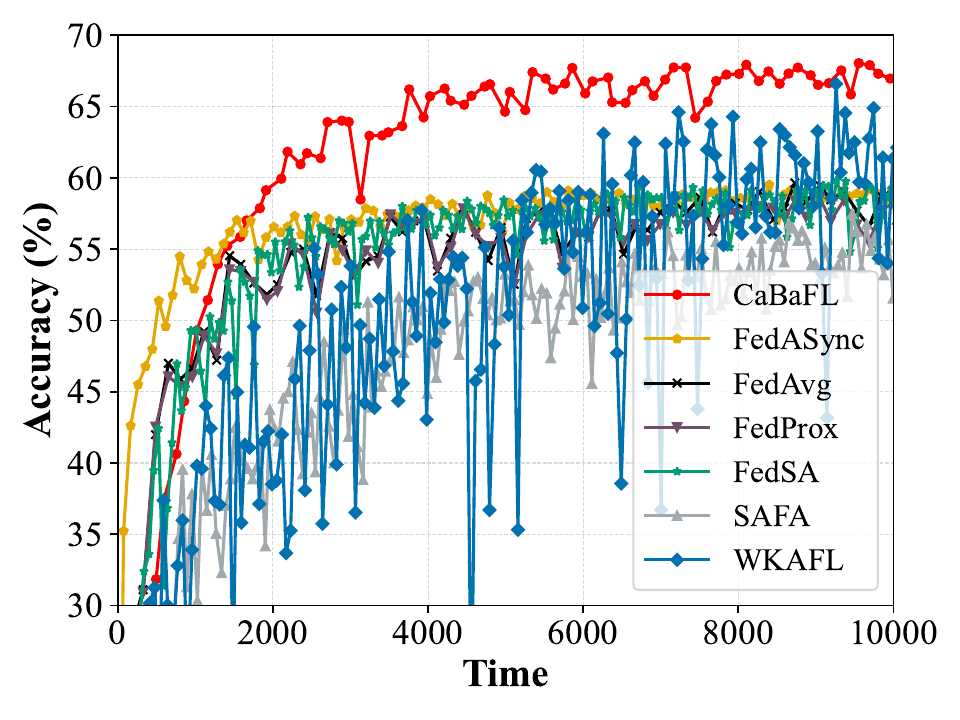}%
\label{}}
\hfil
\subfloat[{\small $20\%$}]{\includegraphics[width=0.4\columnwidth]{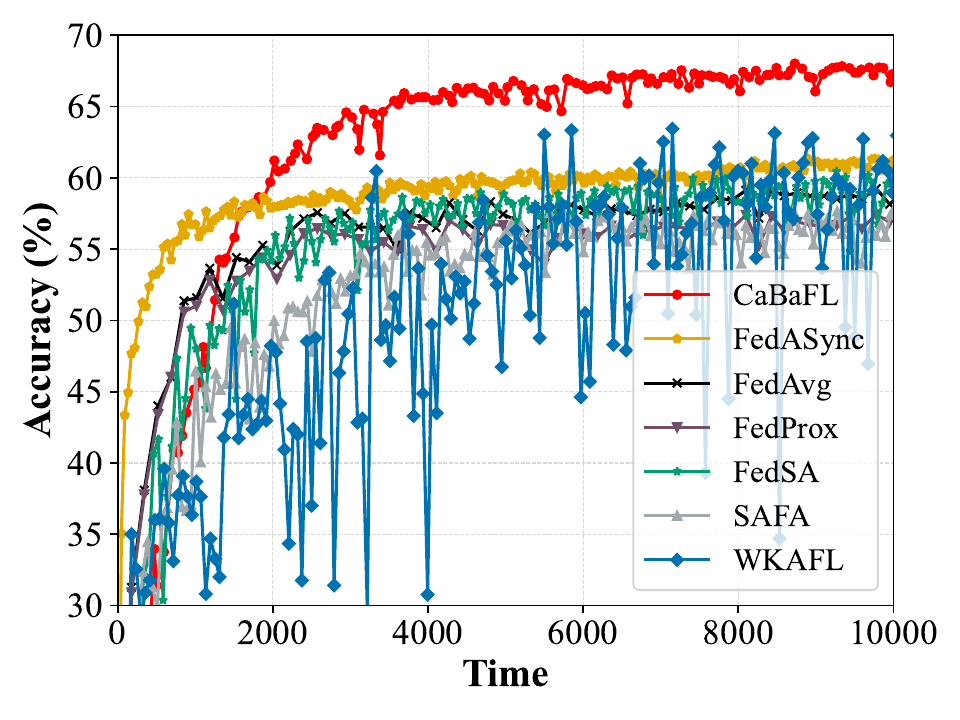}%
\label{}}
\hfil
\subfloat[{\small $50\%$}]{\includegraphics[width=0.4\columnwidth]{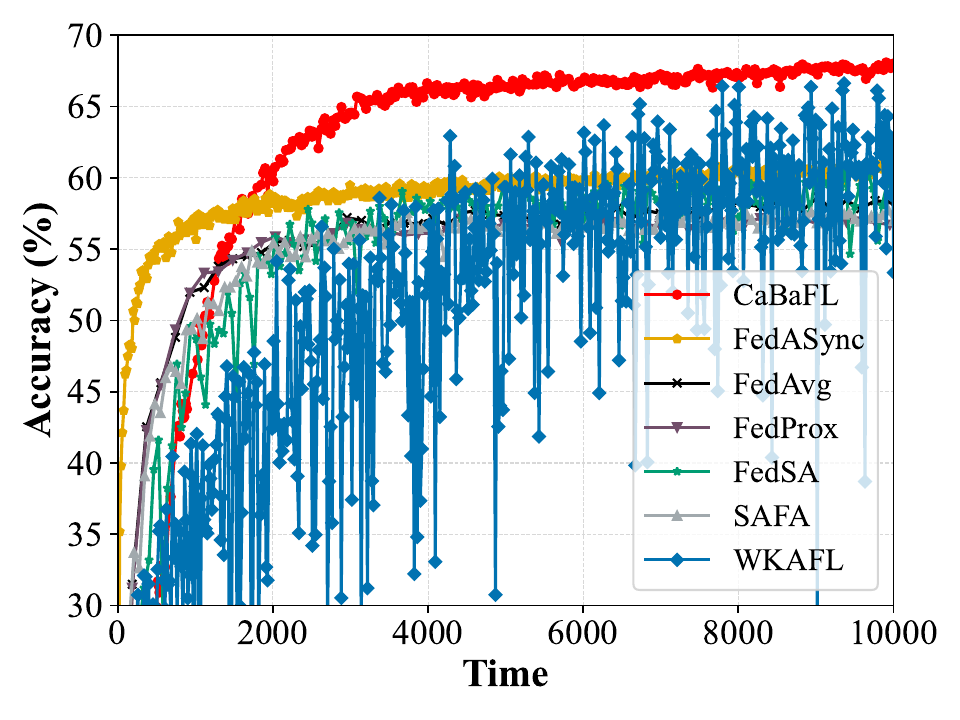}%
\label{}}
\vspace{-0.05 in}
\caption{Learning curves for different ratios of simultaneously training clients.}
\label{fig:fig_extention}
\vspace{-0.15 in}
\end{figure}

By default, we assumed that 10\% of devices were selected to participate in the training simultaneously. 
To investigate the impact of different numbers of devices trained simultaneously, we also considered four different ratios
(i.e., 5\%, 10\%, 20\%, and 50\%) for simultaneously training devices, and conducted experiments using ResNet-18 on CIFAR-10 with $\beta=0.5$.
Note that in the experiments, we did not specify the percentages of 
stragglers using the above ratios. Instead, we assumed that the performance of devices within an experiment follows some Gaussian distribution, where stragglers only denote weak devices with
low training speed.
From 
Figure \ref{fig:fig_extention}, we can find that more selected devices will lead to more stable training convergence. 
This is because more selected devices lead to  more training data, thus alleviating the weight divergence problem during the 
 global model training.

\subsubsection{Impacts of Different Device Performance}
We conducted an experiment to assess the generalization ability of CaBaFL. Our experiment considered the impact of network conditions on delay and the varying computing capabilities of devices on local training time. To facilitate the evaluation, we combined the delay and local training time, assuming that their combinations adhere to Gaussian distributions. We consider five kinds of device conditions, i.e., excellent, high, medium, low, and critical, as illustrated in Table \ref{device_table}. 

\begin{table}[h]
 \vspace{-0.1 in}
\centering
\caption{Training performance for devices.}
\scriptsize
\vspace{-0.1 in}
\label{device_table}
\begin{tabular}{cccccc}
\hline
Quality         & Excellent & High    & Medium  & Low     & Critical \\ \hline
Device Settings & N(10,1)   & N(15,2) & N(20,2) & N(30,3) & N(50,5)  \\ \hline
\end{tabular}
\vspace{-0.1 in}
\end{table}

Based on Table \ref{device_table}, we considered four configurations as shown in Table \ref{device_count}  that have different device compositions to evaluate the generalization ability of CaBaFL.

\begin{table}[h]
\vspace{-0.1 in}
\centering
\caption{Configurations of different device compositions.}
\scriptsize
\label{device_count}
\vspace{-0.1 in}
\begin{tabular}{cccccc}
\hline
\multirow{2}{*}{Config} & \multicolumn{5}{c}{\# of Clients}          \\ \cline{2-6} 
                         & Excellent & High & Medium & Low & Critical \\ \hline
Config1                 & 40        & 30   & 10     & 10  & 10       \\ \hline
Config2                 & 10        & 10   & 10     & 30  & 40       \\ \hline
Config3                 & 10        & 20   & 40     & 20  & 10       \\ \hline
Config4                 & 20        & 20   & 20     & 20  & 20       \\ \hline
\end{tabular}
\vspace{-0.1 in}
\end{table}

We conducted our experiments using the ResNet-18 model on the CIFAR-10 dataset with $\beta=0.1$. Table \ref{device} compares the accuracy between CaBaFL and baselines.

\begin{table}[h]
\vspace{-0.15in}
\centering
\caption{Test accuracy comparison for different device configurations.}
\label{device}
\vspace{-0.1in}
\resizebox{\linewidth}{!}{%
\begin{tabular}{cccccc}
\hline
\multicolumn{2}{c}{Configration} & Config1       & Config2       & Config3       & Config4       \\ \hline
\multirow{7}{*}{\begin{tabular}[c]{@{}c@{}}Test \\ Acuracy\\ (\%)\end{tabular}} & FedAvg & $44.91\pm2.63$ & $44.78\pm2.57$ & $45.45\pm2.83$ & $45.32\pm2.99$ \\ \cline{2-6} 
         & FedProx          & $46.34\pm2.55$ & $44.59\pm2.86$ & $46.14\pm3.11$ & $43.79\pm2.63$ \\ \cline{2-6} 
         & FedASync         & $50.31\pm1.78$ & $48.67\pm1.65$ & $50.12\pm2.17$ & $49.16\pm1.47$ \\ \cline{2-6} 
         & FedSA            & $47.75\pm2.32$ & $46.32\pm1.21$ & $46.51\pm2.54$ & $47.04\pm2.51$ \\ \cline{2-6} 
         & SAFA             & $41.68\pm2.97$ & $40.14\pm2.51$ & $41.32\pm3.02$ & $40.44\pm3.88$ \\ \cline{2-6} 
         & WKAFL            & $49.06\pm6.67$ & $39.30\pm6.65$ & $46.79\pm6.16$ & $39.55\pm5.12$ \\ \cline{2-6} 
         & CaBaFL           & $\textbf{56.96}\pm\textbf{0.58}$ &\textbf{} $\textbf{56.06}\pm\textbf{0.50}$ & $\textbf{56.13}\pm\textbf{0.54}$ & $\textbf{56.74}\pm\textbf{0.60}$ \\ \hline
\end{tabular}%
}
\vspace{-0.1 in}
\end{table}

\begin{table*}[h]
\centering
\vspace{-0.1 in}
\caption{Test accuracy comparison for different task types.}
\vspace{-0.1 in}
\label{tab:task_type}
\begin{tabular}{cccccccc}
\hline
\multirow{2}{*}{Dataset} & \multicolumn{7}{c}{Test Accuracy (\%)}                      \\ \cline{2-8} 
                         & FedAvg & FedProx & FedASync & FedSA & SAFA & WKAFL & CaBaFL \\ \hline
Shakespeare & $51.61\pm0.07$ & $51.63\pm0.12$ & $50.83\pm0.19$ & $51.89\pm0.05$ & $51.13\pm0.23$ & $51.78\pm0.32$ & $\textbf{52.69}\pm\textbf{0.10}$ \\ \hline
Activity    & $95.10\pm0.16$ & $95.23\pm0.12$ & $95.06\pm0.08$ & $95.27\pm0.09$ & $93.47\pm0.45$ & $95.17\pm0.76$ & $\textbf{95.43}\pm\textbf{0.23}$ \\ \hline
\end{tabular}%
\vspace{-0.25in}
\end{table*}

From this table, we can find that CaBaFL can achieve the highest and stablest accuracy in all four cases. 
Meanwhile, we can observe
that the accuracy of baselines decreases significantly when the numbers of stragglers increase, while the accuracy differences of our approach are small. This is mainly because both synchronous and semi-asynchronous methods must wait for a certain number of models before aggregating. Therefore, their accuracy is affected by stragglers. Moreover, the number of stale models increased due to the increase in the number of stragglers, leading to a decrease in accuracy.

\subsubsection{Impacts of Different Model Training Times}
\begin{figure}[htp]
  \vspace{-0.1 in}
  \begin{center} 
		\includegraphics[width=0.25\textwidth]{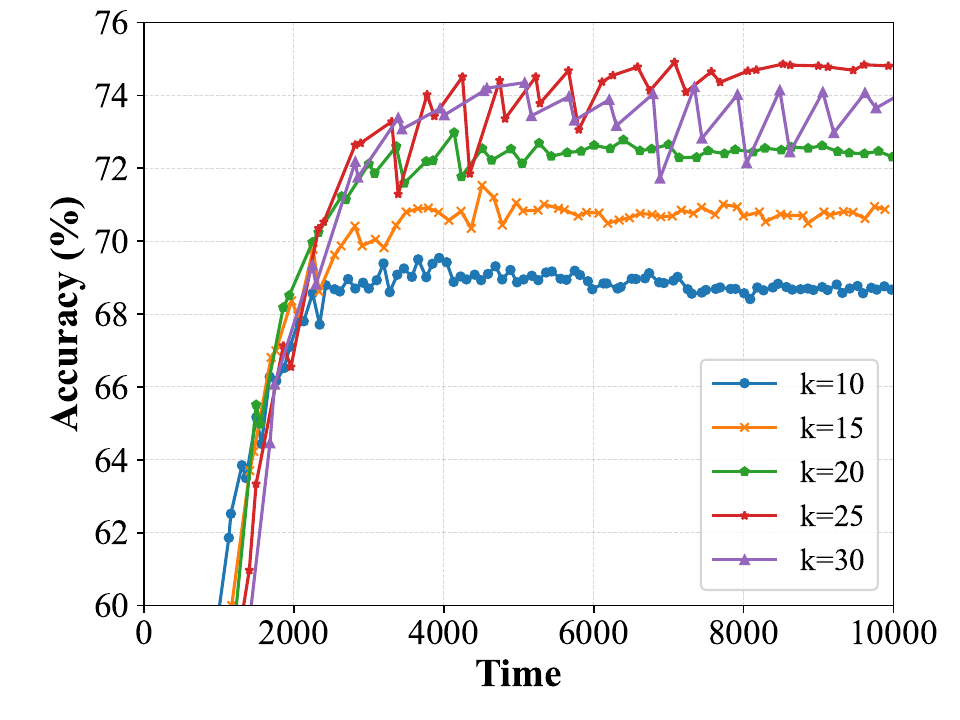}
  \vspace{-0.1 in}
		\caption{Impact of training times on test accuracy.}
		\label{fig:diff_k} 
	\end{center}
 \vspace{-0.2in}
\end{figure}

To investigate the impact of model training times $k$, we set up five different model training times, respectively 10, 15, 20, 25, and 30,  and conducted experiments using ResNet-18 model on CIFAR-10 dataset with IID distribution. 
Figure \ref{fig:diff_k} exhibits the experiment result. We can find that as the model training times $k$ increase, the accuracy of the global model improves, but too high a number of model training times leads to a decrease in global model accuracy and instability of the learning curve. 

\subsubsection{Impacts of Task Types}
To evaluate
the generalization ability of our approach to different types of tasks, we conducted experiments on two well-known non-image-based datasets,  i.e., the text type dataset Shakespeare~\cite{femnist} and 
the table type dataset Activity~\cite{activity}, using the LSTM model and MLP model, respectively.  
From Table \ref{tab:task_type}, we can find that CaBaFL can achieve the best results in all two cases, indicating the generalization ability of CaBaFL on different tasks.


\begin{figure}[h]
\vspace{-0.25 in}
\centering
\footnotesize
\subfloat[{\small $\beta=0.1$}]{\includegraphics[width=0.4\columnwidth]{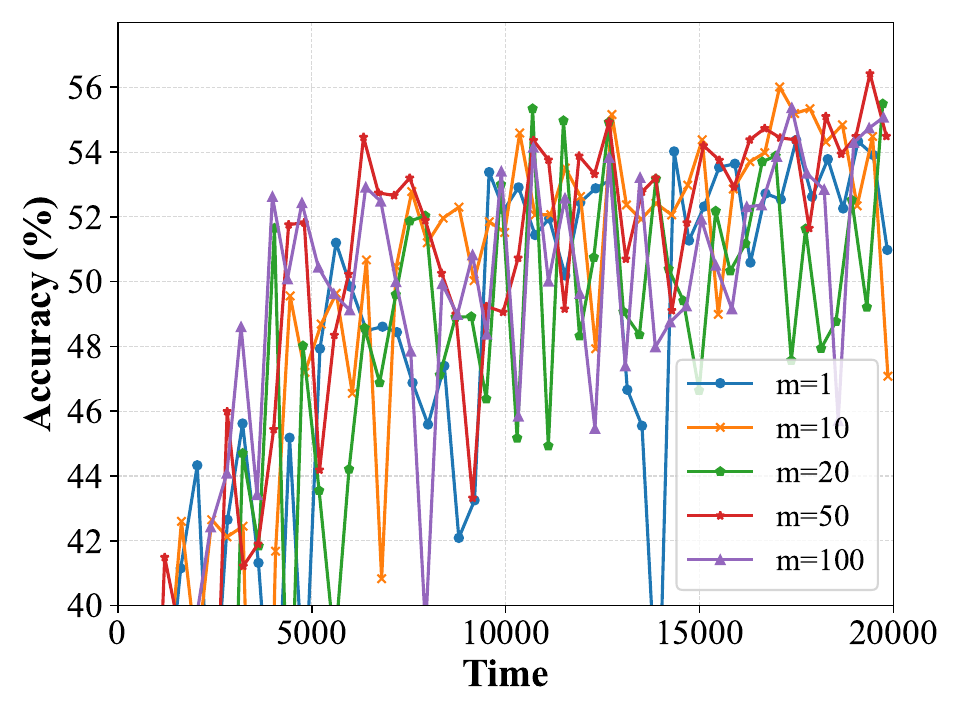}%
\label{}}
\hfil
\subfloat[{\small IID}]{\includegraphics[width=0.4\columnwidth]{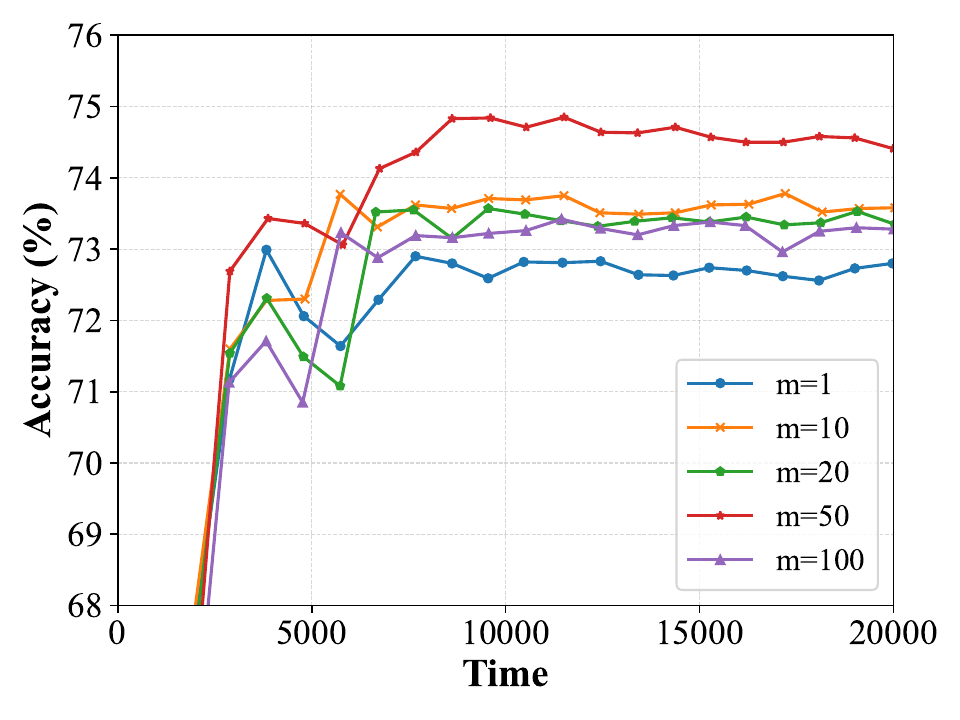}%
\label{}}
\vspace{-0.1 in}
\caption{Impact of the feature collection cycle.}
\label{fig:diff_m}
\vspace{-0.1 in}
\end{figure}

\subsubsection{Impacts of  Feature Collection Cycles} To investigate the impact of feature collection cycles, We assume that feature collection is performed after every $m$ cache aggregation.  We set $m$ to be 1, 10, 20, 50, and 100, respectively, and conduct experiments on CIFAR-10 dataset and ResNet-18 model with both the IID distribution and the Dirichlet distribution (with $\beta=0.1$). Figure \ref{fig:diff_m} exhibits the experiment results. From the results, we can find that both too-fast and too-slow feature collection cycles result in lower global model accuracy. In addition, a feature collection cycle that is too fast will significantly increase the communication overhead because the server needs to frequently send the global model to all devices for feature collection. 

\subsection{Ablation Study}
\subsubsection{Key Components of CaBaFL}
To demonstrate the effectiveness of CaBaFL, we investigated five variants of CaBaFL: i) {\it Conf1} represents selecting devices based only on the similarity between model feature and global feature; ii) {\it Conf2}  denotes selecting devices based only on data size; iii) {\it Conf3} indicates randomly selecting devices; iv) {\it Conf4} represents aggregating models in the L2 Cache, which means only L2 Cache is available in the server; and v) {\it Conf5} denotes averaging aggregation during model aggregation. 

\begin{table}[h]
\vspace{-0.1in}
\centering
\caption{Comparison of test accuracy among CaBaFL and its variants.}
\label{abliation_table}
\vspace{-0.1in}
\resizebox{0.99\linewidth}{!}{%
\begin{tabular}{ccccc}
\hline
\multirow{2}{*}{Settings} & \multicolumn{4}{c}{Test Accuracy(\%)}                             \\ \cline{2-5} 
                          & $\beta=0.1$        & $\beta=0.5$        & $\beta=1.0$        & $IID$  \\ \hline
CaBaFL                    & $55.46\pm0.67$ & $69.31\pm0.18$ & $72.84\pm0.36$ & $74.83\pm0.07$ \\ \hline
Conf1                     & $52.54\pm0.95$ & $69.05\pm0.23$ & $72.73\pm0.23$ & $73.99\pm0.18$ \\ \hline
Conf2                     & $54.16\pm0.65$ & $68.04\pm0.20$ & $71.59\pm0.35$ & $72.60\pm0.18$ \\ \hline
Conf3                     & $53.50\pm1.23$ & $69.16\pm0.19$ & $72.80\pm0.34$ & $72.67\pm0.26$ \\ \hline
Conf4                     & $53.29\pm2.12$ & $69.27\pm0.42$ & $72.49\pm0.26$ & $73.91\pm0.09$ \\ \hline
Conf5                     & $53.08\pm0.49$ & $68.30\pm0.25$ & $71.47\pm0.12$ & $72.99\pm0.18$ \\ \hline
\end{tabular}%
}
\vspace{-0.1in}
\end{table}

\begin{figure}[h]
\vspace{-0.25 in}
\centering
\footnotesize
\subfloat[{\small $\beta=0.1$}]{\includegraphics[width=0.4\columnwidth]{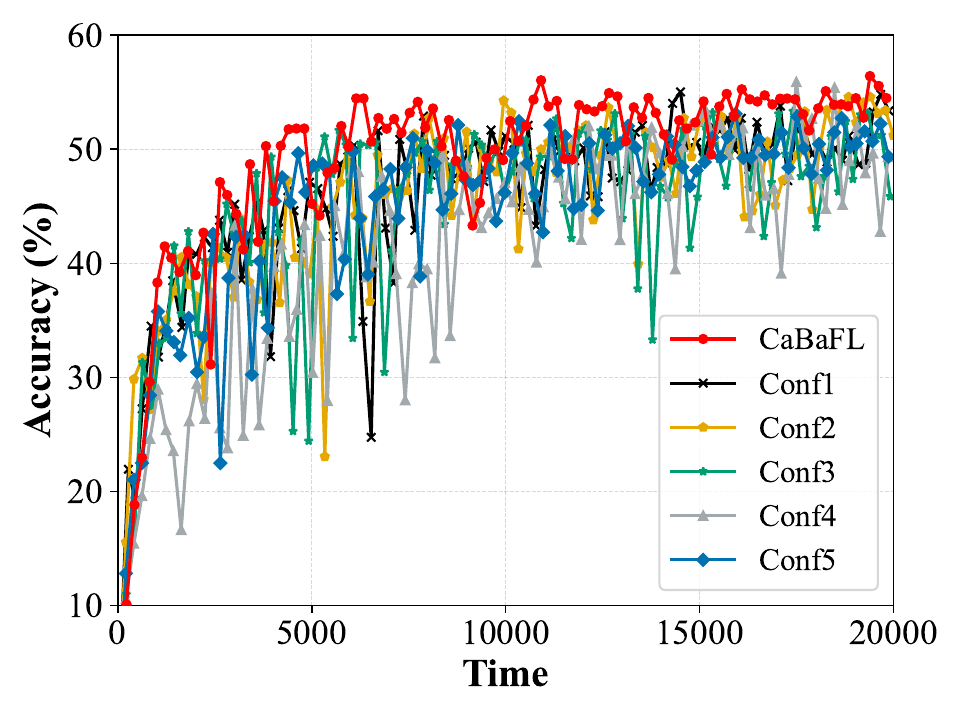}%
\label{fig:real_testbed}}
\hfil
\subfloat[{\small $\beta=0.5$}]{\includegraphics[width=0.4\columnwidth]{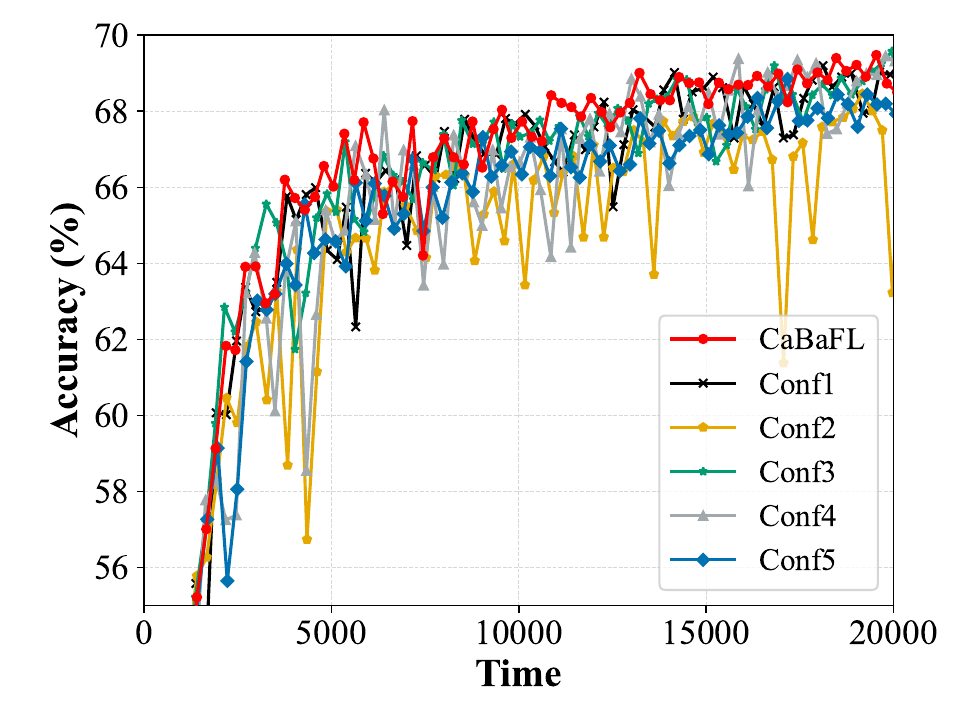}%
\label{fig:real-cifar-10-IID-ls}}
\hfil
\subfloat[{\small $\beta=1.0$}]{\includegraphics[width=0.4\columnwidth]{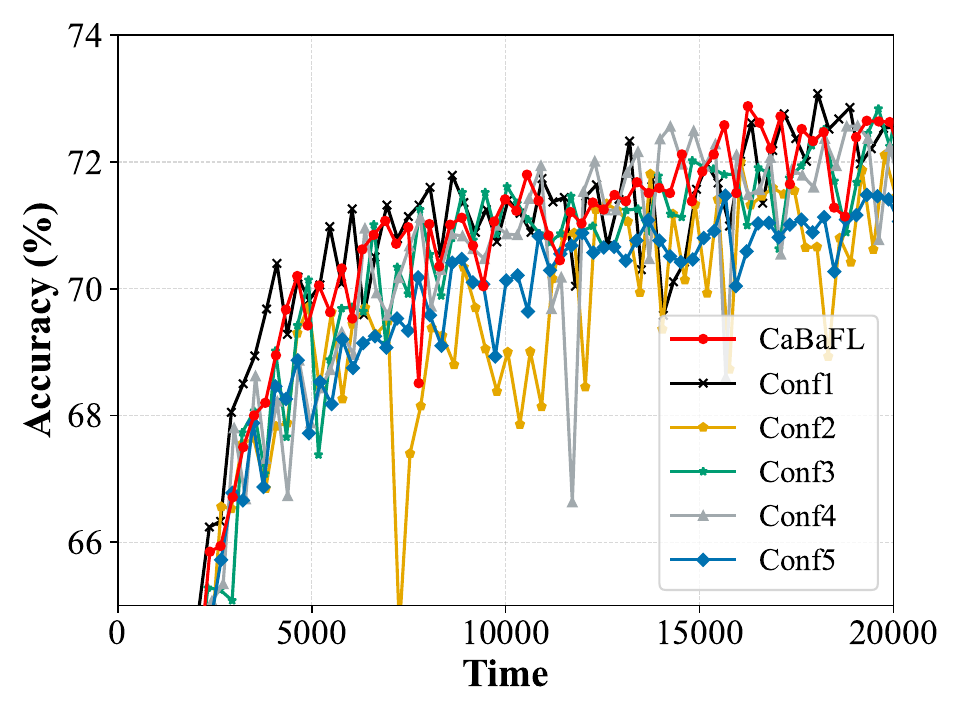}%
\label{fig:real-cifar-10-IID-lm}}
\hfil
\subfloat[{\small IID}]{\includegraphics[width=0.4\columnwidth]{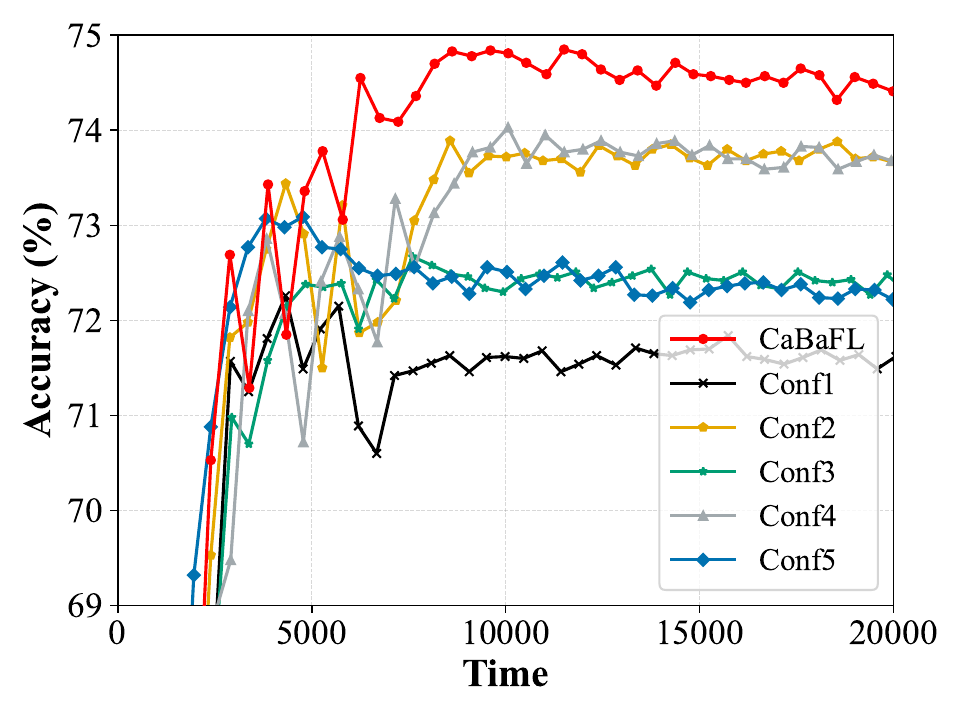}%
\label{fig:real_testbed}}
\vspace{-0.1in}
\caption{Ablation study results.}
\label{fig:fig_abliation}
\vspace{-0.1 in}
\end{figure}

We conducted experiments using the ResNet-18 model on CIFAR-10 with both non-IID and IID settings, and the results are presented in Table \ref{abliation_table} and Figure \ref{fig:fig_abliation}. Table \ref{abliation_table} shows that CaBaFL achieves the highest test accuracy among all six designs.
By comparing the performance of {\it Conf4} and CaBaFL, we can observe that our 2-level cache structure significantly improves the stability of model training and enhances model accuracy in scenarios with high levels of data heterogeneity. By comparing the performance of {\it Conf5} and CaBaFL, we observed that our cache aggregation strategy significantly improves the model accuracy. Furthermore, by comparing the performance of {\it Conf1} to {\it Conf3} with that of CaBaFL, we observed that our feature balance-guided device selection strategy is effective in different scenarios of data heterogeneity. Especially under the extreme data heterogeneity condition where $\beta=0.1$, our device selection strategy can significantly improve the accuracy of the global model compared to {\it Conf3}, which is random device selection.

\subsubsection{Activation Amounts versus Hard Labels}
\begin{figure}[h]
\vspace{-0.2 in}
\centering
\footnotesize
\subfloat[{\scriptsize CIFAR-10, $\beta=0.1$}]{\includegraphics[width=0.32\columnwidth]{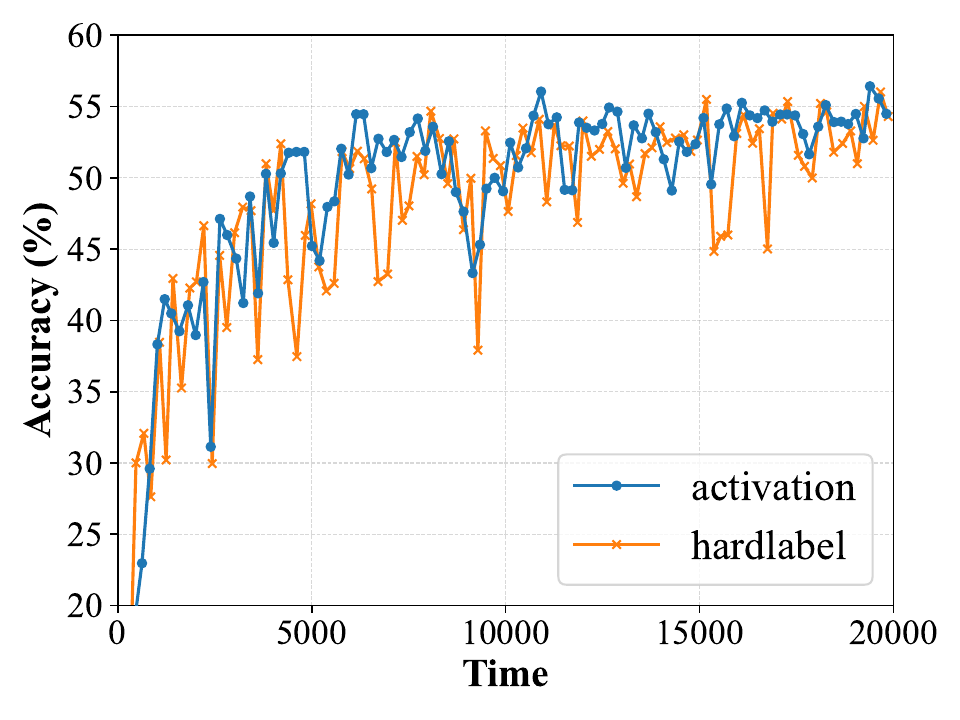}%
\label{}}
\hfil
\subfloat[{\scriptsize CIFAR-10, IID}]{\includegraphics[width=0.32\columnwidth]{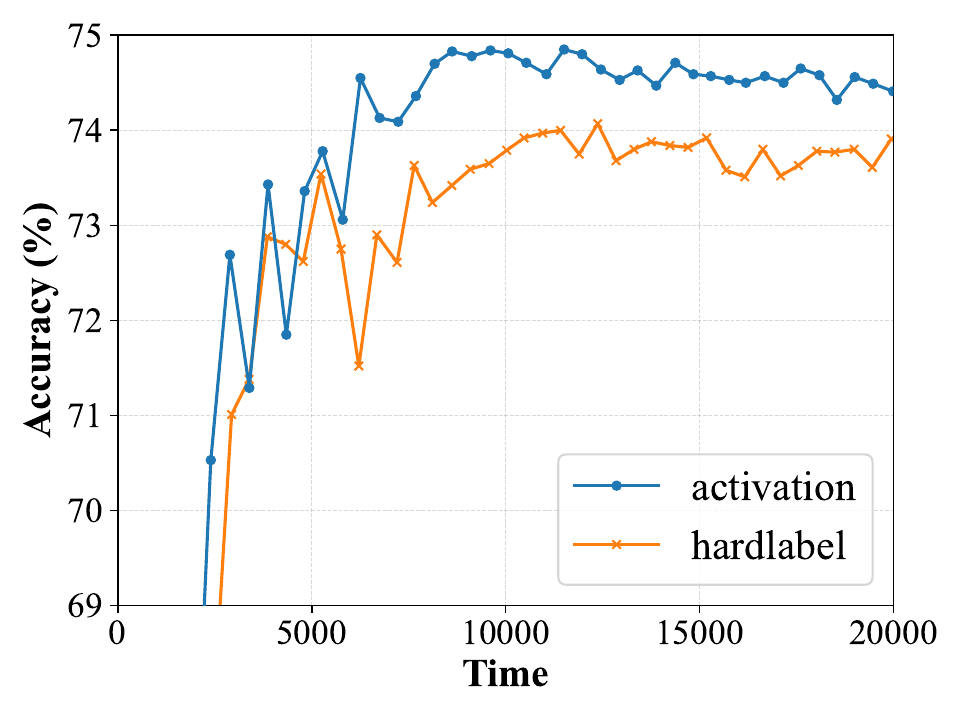}%
\label{fig:real-cifar-10-IID-ls}}
\hfil
\subfloat[{\scriptsize CIFAR-100, Special }]{\includegraphics[width=0.32\columnwidth]{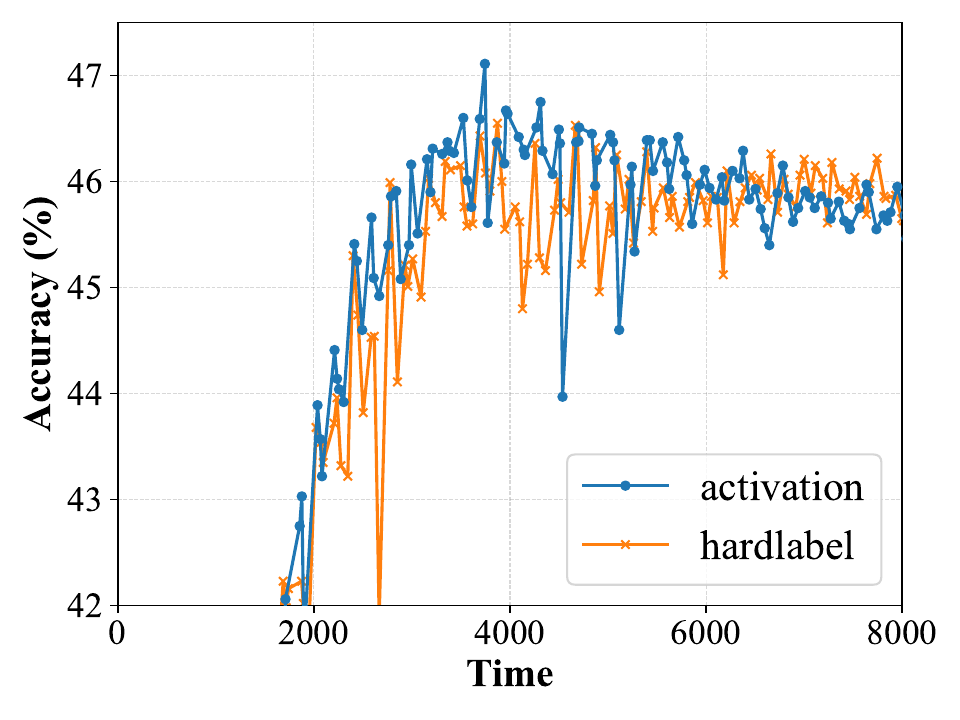}%
\label{fig:real-cifar-10-IID-lm}
\vspace{-0.1in}}
\caption{Comparison of accuracy based on activation amounts and hard labels.}
\label{fig:act_VS_hd}
\vspace{-0.25in}
\end{figure}

We evaluated the impact of activation amounts and hard labels on the 
training performance of CaBaFL. 
Figure \ref{fig:act_VS_hd}(a) and (b) show the results conducted on the combination of CIFAR-10 and ResNet-18 within IID and non-IID ($\beta=0.1$), respectively. 
We can find activation amount-based CaBaFL outperforms hard label-based CaBaFL, since activation distributions provide a fine-grained
representation for data distributions. 
Note that the CIFAR-100 dataset consists of 20 coarse-grained categories, which can be refined into 100 fine-grained categories.
To reproduce Observation 2 in Section \ref{sec:motivation}, we construct a special data distribution for devices based on CIFAR-100, where the device data are IID according to coarse-grained categories but non-IID according to fine-grained categories.
Figure \ref{fig:act_VS_hd}(c) shows the training processes using ResNet-18, where 
activation amount-based CaBaFL outperforms hard label-based CaBaFL
due to its fine-grained representation of data distributions. 

\subsection{Real Test-bed Evaluation}
\begin{figure}[h]
\vspace{-0.15 in}
  \begin{center} 
		\includegraphics[width=0.25\textwidth]{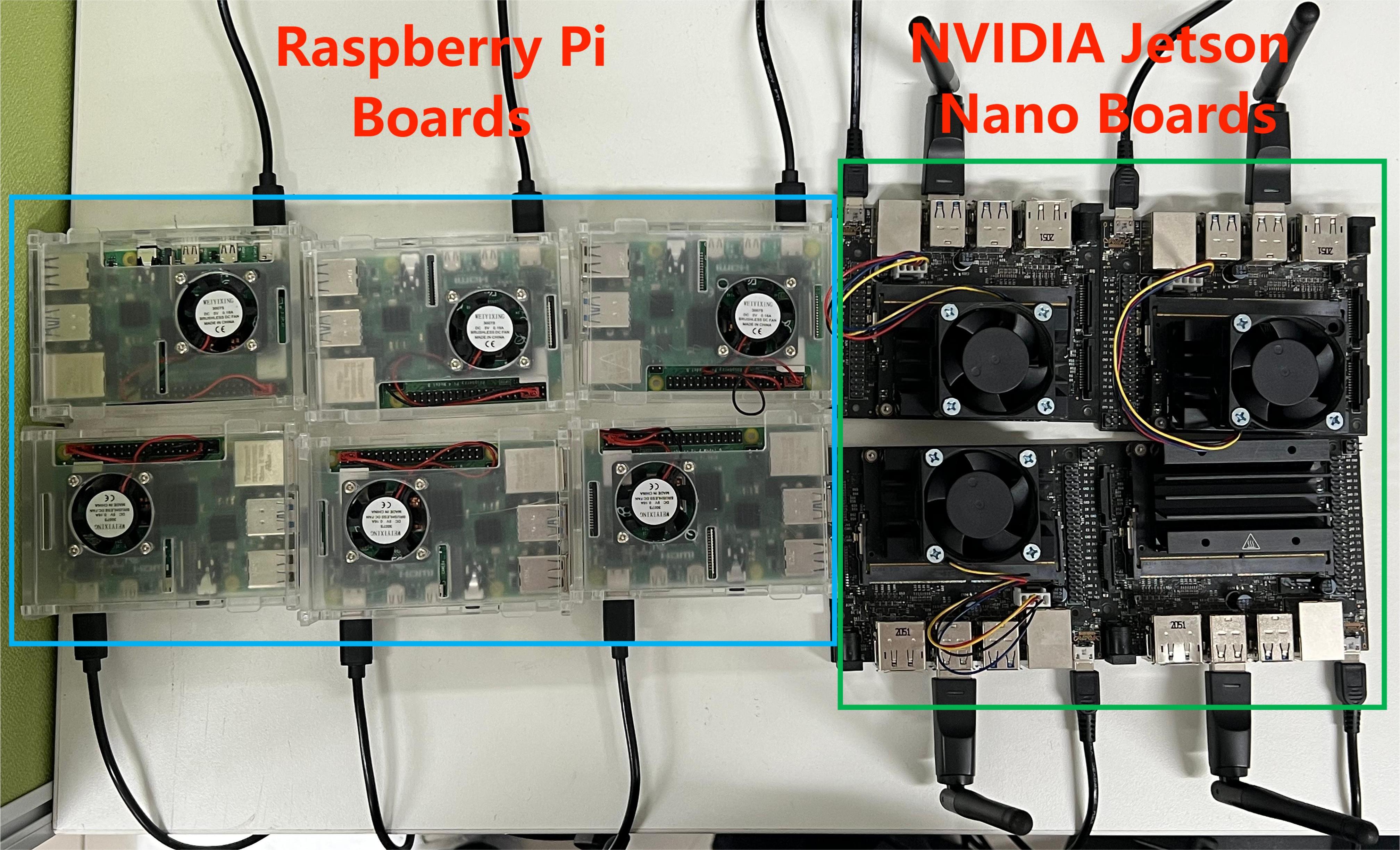}
  \vspace{-0.1 in}		
  \caption{Our real test-bed platform.}
		\label{fig:real_devices} 
	\end{center}
 \vspace{-0.15 in}
\end{figure}

To evaluate the effectiveness of our approach in practical scenarios, we conducted experiments on real devices using CNN models and the CIFAR-10 dataset, considering both IID and non-IID ($\beta=1.0$) scenarios. Figure \ref{fig:real_devices} shows our test-bed platform, which uses: i) an Ubuntu-based cloud server equipped with an Intel i9 CPU, 32GB memory, and an NVIDIA RTX 3090Ti GPU, and ii)  four NVIDIA Jetson Nano boards and six Raspberry Pi boards as heterogeneous clients.

\begin{figure}[htp]
\centering
\footnotesize
\subfloat[{\scriptsize Wall Clock Time}]
{\includegraphics[width=0.32\columnwidth]{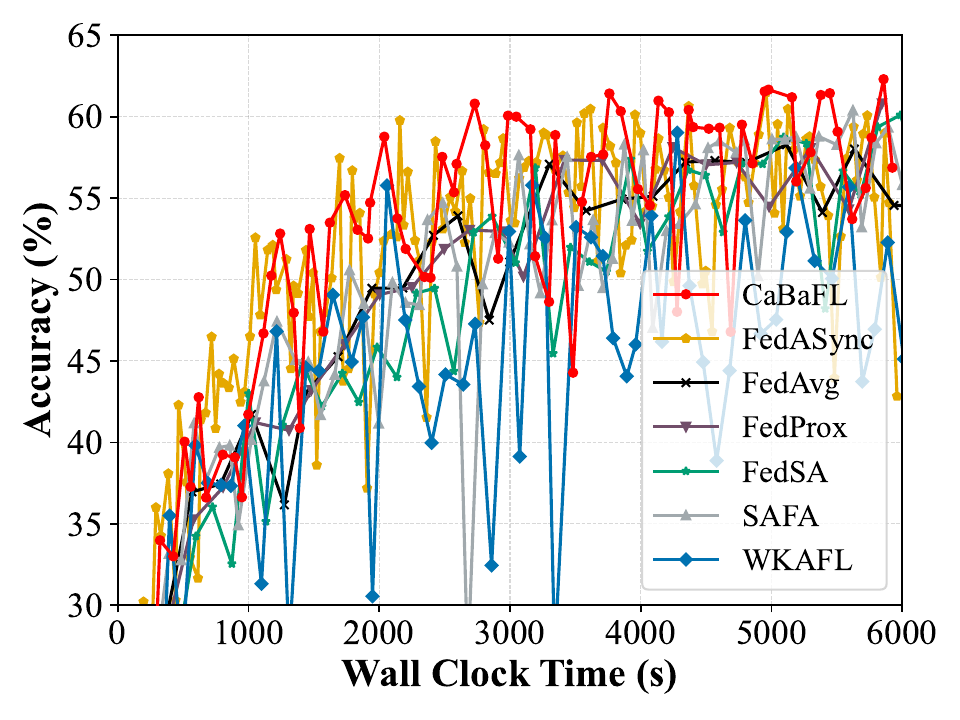}%
\label{fig:real_1.0}}
\hfil
\subfloat[{\scriptsize Commu. Overhead}]
{\includegraphics[width=0.32\columnwidth]{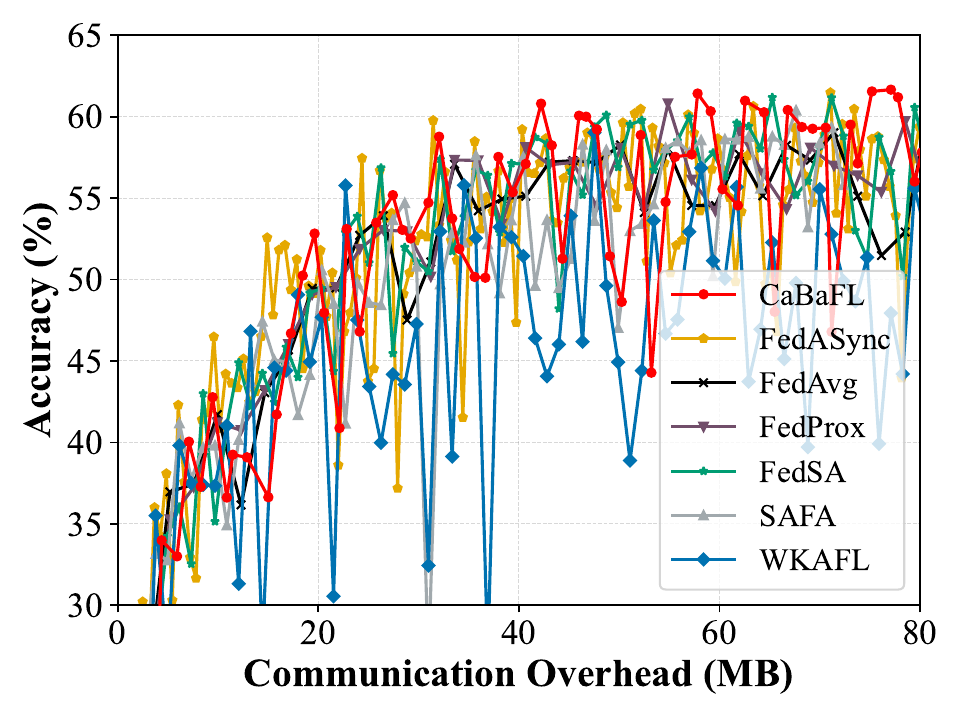}%
\label{fig:real-cifar-10-IID-ls}}
\hfil
\subfloat[{\scriptsize Energy Consumption}]
{\includegraphics[width=0.32\columnwidth]{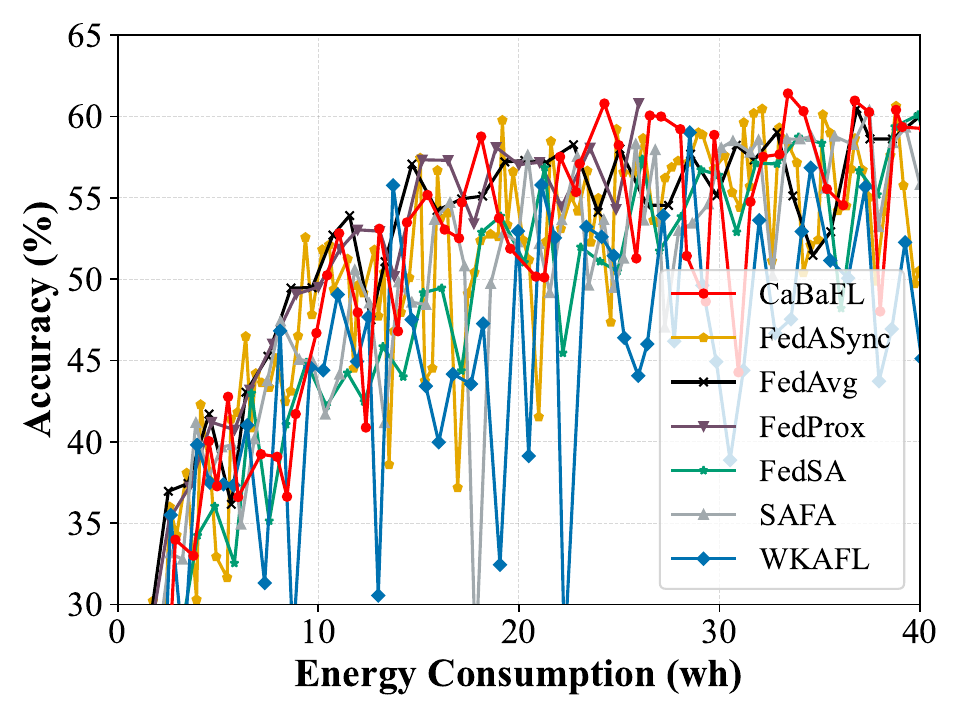}%
\label{fig:real-cifar-10-IID-ls}}
                        \vspace{-0.1 in}
\caption{Comparison of accuracy for different system metrics ($\beta=1.0$).}
\vspace{-0.25 in}
\label{fig:real_devices_1.0}
\end{figure}

\begin{figure}[htp]
\vspace{-0.1 in}
\centering
\footnotesize
\subfloat[{\scriptsize Wall Clock Time}]
{\includegraphics[width=0.32\columnwidth]{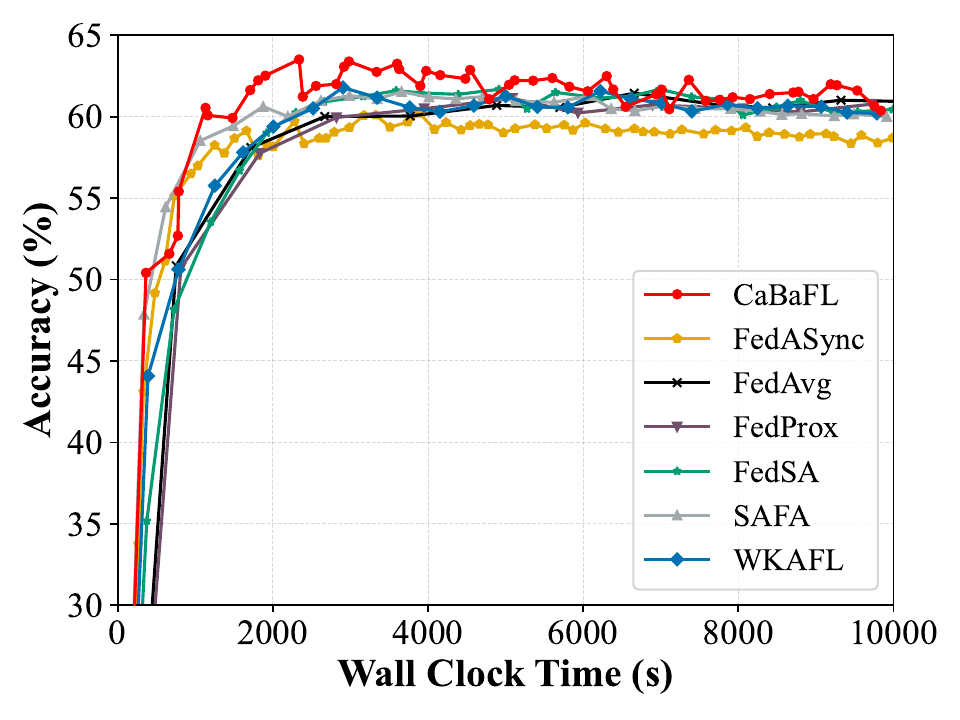}%
\label{fig:real_1.0}}
\hfil
\subfloat[{\scriptsize Commu. Overhead}]
{\includegraphics[width=0.32\columnwidth]{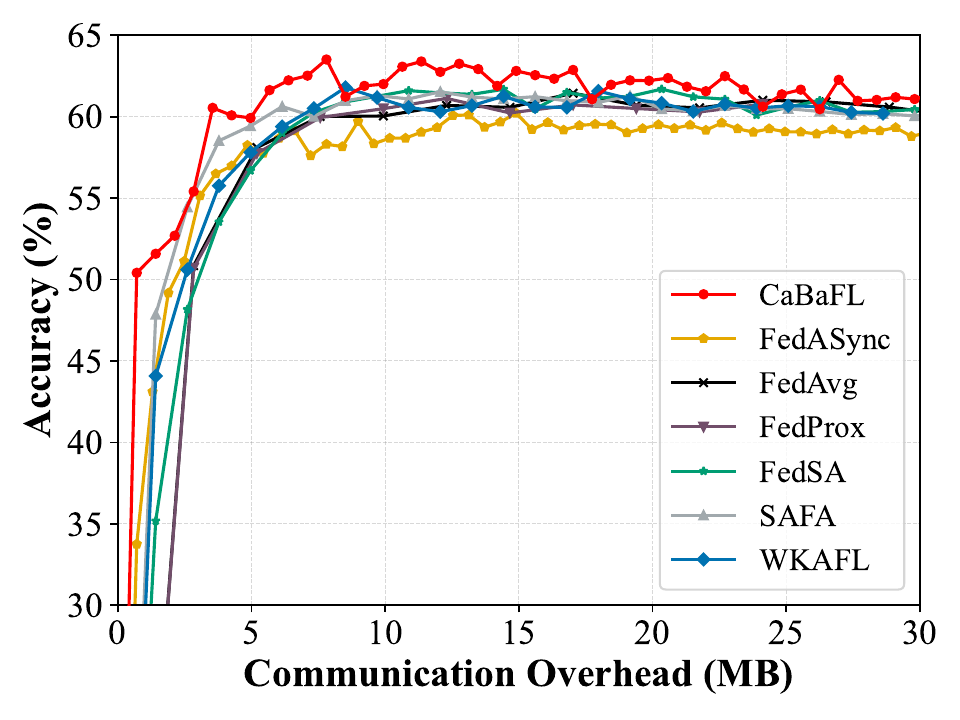}%
\label{fig:real-cifar-10-IID-ls}}
\hfil
\subfloat[{\scriptsize Energy Consumption}]
{\includegraphics[width=0.32\columnwidth]{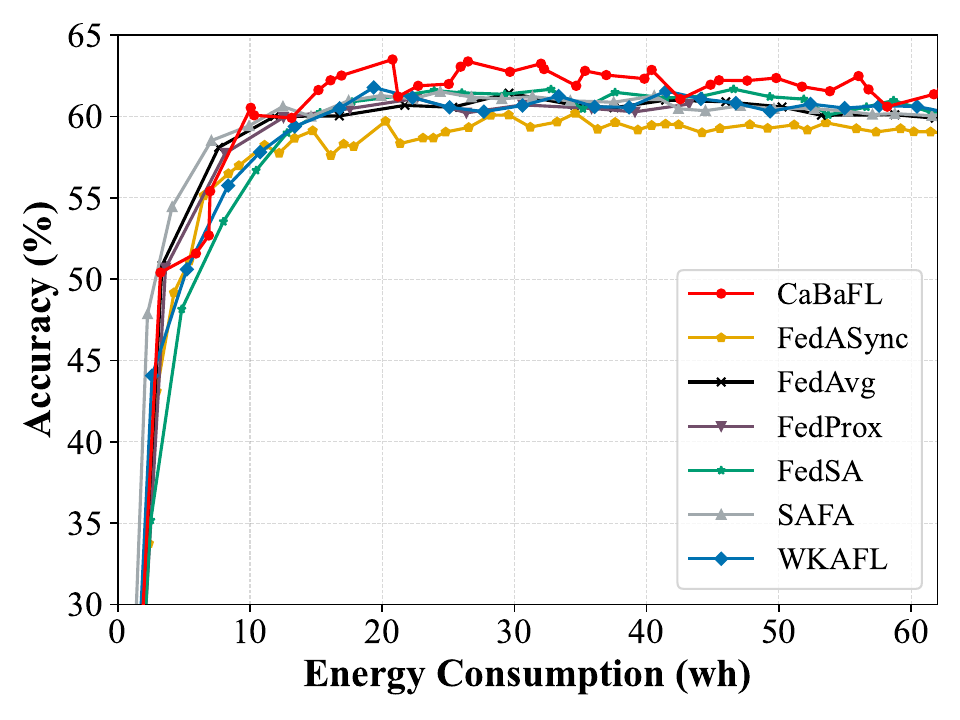}%
\label{fig:real-cifar-10-IID-ls}}
\vspace{-0.1 in}
\caption{Comparison of accuracy for different system metrics (IID).}
 \vspace{-0.25in}
\label{fig:real_devices_iid}
\end{figure}

Figures~\ref{fig:real_devices_1.0} and \ref{fig:real_devices_iid} present a comparison of accuracy for different system metrics on our test-bed platform.  
 We can find that CaBaFL can also achieve the best performance compared to all the baselines. Please note that the number of samples per device in the real test-bed platform is much bigger than the number of samples per device in the simulation experiment. As a result, the experiment in Figure 11 is more likely to cause much more severe catastrophic forgetting even when $\beta=1.0$, resulting in large amplitudes of the learning curves. Note that since the local epoch is smaller with $\beta=1.0$ than with IID distribution, the communication overhead is greater with $\beta=1.0$ than with IID distribution. 

\subsection{Impact of Hyperparameters}
To evaluate the impact of hyperparameters in CaBaFL, we set different configurations for  $\alpha$, $\gamma$, and $\sigma$, respectively, and performed experiments in the CIFAR-10 dataset using ResNet-18 model with IID and Non-IID ($\beta=0.1$) scenarios.
Tables \ref{tab:alpha}-\ref{tab:gamma} exhibit the experiment results.

Table \ref{tab:alpha} shows the inference accuracy with different settings of $\alpha$.
From Table \ref{tab:alpha}, we can find that when the non-IID degree is high, a smaller $\alpha$ can improve the accuracy of the model, while under the IID distribution, a larger $\alpha$ can improve the accuracy of the model. This is mainly because when the degree of non-IID is high, the local data size of the device varies greatly, so a smaller $\alpha$ is needed to balance this difference, while the opposite is true in the case of IID.

\begin{table}[h]
\centering
\vspace{-0.15in}
\caption{Impact of $\alpha$ on test accuracy.}
\vspace{-0.1 in}
\footnotesize
\resizebox{\linewidth}{!}{%
\begin{tabular}{ccccc}
\hline
\multirow{2}{*}{\begin{tabular}[c]{@{}c@{}}Hetero.\\ Settings\end{tabular}} & \multicolumn{4}{c}{Value of $\alpha$} \\ \cline{2-5} 
            & 0.5            & 0.7            & 0.9            & 1              \\ \hline
$\beta=0.1$ & $55.46\pm0.67$ & $54.63\pm0.46$ & $55.14\pm0.79$ & $53.70\pm0.79$ \\ \hline
$IID$       & $73.07\pm0.09$ & $73.31\pm0.11$ & $73.48\pm0.20$ & $74.83\pm0.07$ \\ \hline
\end{tabular}
}
\label{tab:alpha}
\vspace{-0.1 in}
\end{table}

Table \ref{tab:sigma} shows the impact of $\sigma$. We can find that when the non-IID degree is high, a larger $\sigma$ can improve the accuracy of the model, while under the IID distribution, a smaller $\sigma$ can improve the accuracy of the model. Please note that CaBaFL normalizes the number of times a device is selected before calculating the variance of the number of times the device is selected, so the value of $\sigma$ is less than 0.

\begin{table}[h]
\centering
\vspace{-0.15 in}
\caption{Impact of $\sigma$ on test accuracy.}
\vspace{-0.1 in}
\footnotesize
\resizebox{\linewidth}{!}{%
\begin{tabular}{ccccc}
\hline
\multirow{2}{*}{\begin{tabular}[c]{@{}c@{}}Hetero.\\ Settings\end{tabular}} & \multicolumn{4}{c}{Value of $\sigma$} \\ \cline{2-5} 
            & $1\times 10^{-6}$ & $2\times 10^{-6}$ & $3\times 10^{-6}$ & $4\times 10^{-6}$ \\ \hline
$\beta=0.1$ & $55.19\pm0.27$    & $53.60\pm0.33$    & $55.46\pm0.67$    & $54.44\pm1.07$    \\ \hline
$IID$       & $74.83\pm0.07$    & $72.80\pm0.10$    & $73.17\pm0.15$    & $73.72\pm0.09$    \\ \hline
\end{tabular}
}
\label{tab:sigma}
\vspace{-0.15 in}
\end{table} 

Table \ref{tab:gamma} shows the impact of $\gamma$. We can find that when the value of $\gamma$ is appropriate, CaBaFL can achieve better global model accuracy. When the non-IID degree is high, a larger $\gamma$, i.e., filtering more models to make the cumulative training data of the model in the L1 cache more balanced, is required to achieve higher global model accuracy, while under the IID distribution, a smaller $\gamma$, i.e., filtering few models, allows CaBaFL to achieve higher model accuracy.

\begin{table}[h]
\vspace{-0.15in}
\centering
\caption{Impact of $\gamma$ on test accuracy.}
\vspace{-0.1in}
\footnotesize
\resizebox{\linewidth}{!}{%
\begin{tabular}{ccccccc}
\hline
\multirow{2}{*}{\begin{tabular}[c]{@{}c@{}}Hetero.\\ Settings\end{tabular}} &
  \multicolumn{4}{c}{Value of $\gamma$} \\ \cline{2-5} 
 & 0.4 & 0.3 &  0.2 & 0.01 \\ \hline
$\beta=0.1$ &
  $54.59\pm1.62$ &
  $55.56\pm0.77$ &
  $54.08\pm1.04$ &
  $53.26\pm1.95$ \\ \hline
$IID$ &
  $73.19\pm0.10$ &
  $73.06\pm0.14$ &
  $73.11\pm0.11$ &
  $74.83\pm0.07$ \\ \hline
\end{tabular}%
}
\label{tab:gamma}
\vspace{-0.15in}
\end{table}

\section{Discussion}
\label{discussion}
\subsection{Fairness and Workload Balance}
\label{sec:6.a}
\begin{figure}[h]
\vspace{-0.15in}
  \begin{center} 
		\includegraphics[width=0.3\textwidth]{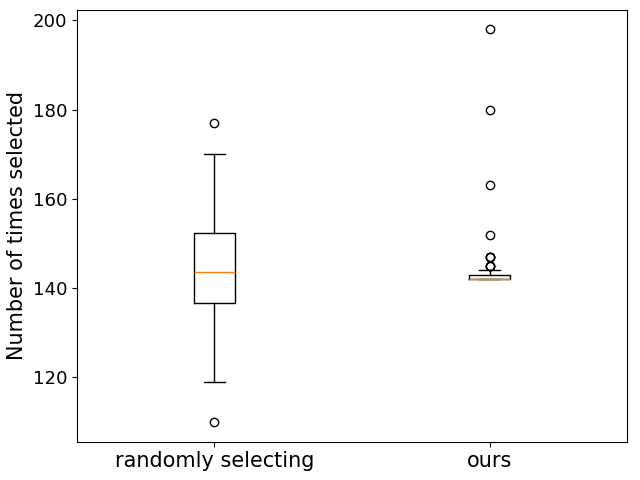}
\vspace{-0.1in}		
  \caption{Comparison between random selection strategy and our feature balance-guided device selection strategy.}
  \vspace{-0.35 in}
		\label{fairness} 
	\end{center}
\end{figure}

 We strive to address the fairness and workload balance issues in our device selection mechanism, where
 we use the hyperparameter $\sigma$ to control the fairness of device selection (see Line 3 of Algorithm 2). 
 During the FL training, if the variance of the number of device selections exceeds $\sigma$, CaBaFL will select idle devices that have been touched least frequently, ensuring the fairness of device selection. We recorded the number of times a device was selected using our device selection strategy as well as the random selection strategy. From   Figure \ref{fairness}, we can find that compared to the random selection strategy, except for a few outliers, the number of selected devices for our strategy is very close, demonstrating that the workload between devices is balanced.

\subsection{Complexity and Scalability}
\label{6.b}
Assume that the dimension size of flattened feature distributions is $d$, the current FL training round index is  $k$, and the number of activated devices in a round is $n$.
Let $D$ be the number of all devices, and $m$ be 
the total number of model parameters. The memory and time complexity analysis of our approach is as follows.

\textbf{Time Complexity.} The time complexity of our method mainly comes from Cache Update and Device Selection. i) Cache Update:  From Lines 9-12 of Algorithm 
 \ref{cabafl}, we can figure out that the time complexity of Cache Update is O($k\log k+d$). ii) Device Selection: From Lines 10-16 of Algorithm \ref{selectdevice}, we can determine the time complexity of Device Selection is O($n(d+n)$). Above all, the overall time complexity of our method is O($k\log k+d+n(d+n)$). Even for a FL system with 1000 devices, one round of Cache Update and Device Selection costs less than 2ms, which is negligible compared with local training time.

\textbf{Memory Complexity.} The memory complexity of our method mainly also comes from Cache Update and Device Selection. i) Cache Update: Although there are two caches in our method, we can only maintain the L1 cache, and the L2 cache can only exist logically. Because the model in the L2 cache can be deleted from memory after being sent to the device for training. Therefore, the memory complexity is O($nm$). In addition, our method also maintains the model feature distribution and device feature distribution, and its memory complexity is O($(D+n)d$). ii) Device Selection: From Algorithm~\ref{selectdevice}, we can determine that the memory complexity of device selection is O($Dd+n$). Since $nm\gg Dd+dn+n$, the overall memory complexity of our method is approximately O($nm$), which is the same as the traditional FL.



\vspace{-0.15in}
\subsection{Security Risk and Limitation}
In the FL training process of CaBaFL, an intermediate model continuously traverses multiple devices to perform local training before the next aggregation operation. It means that a model locally trained by one device can be accessed by its successive counterpart. If an adversary is assigned such a trained model and knows its predecessor, there could be a risk of privacy leaks. However, this situation is difficult to occur in our approach, since the 
server controls the device traversing process for a specific intermediate model by using our proposed device selection strategy. 
 In this way,  CaBaFL does not allow  adversaries to infer its
 predecessor devices, thus mitigating the risk of privacy leaks. 
%
%
Note that it is also hard for CaBaFL to 
directly adopt secure aggregation for privacy protection. Although various privacy-preserving mechanisms (e.g., the variational Bayes-based method) can be easily integrated into CaBaFL to secure clients, they may lead to a slight decrease in performance. 
Therefore, we plan to design a better secure aggregation method without deteriorating inference accuracy in our future work.
Moreover, since CaBaFL involves multiple hyperparameters,  how to quickly figure out an optimal combination for them to explore the greatest collaborative learning potential among devices is another interesting direction that is worthy of further study. 


\section{Conclusion}
\label{sec:conclusion}

To improve the inference performance of Federated Learning (FL) in large-scale AIoT applications, this paper introduced a new asynchronous FL method named CaBaFL, which can effectively mitigate the notorious straggler and data imbalance problems caused by device and data heterogeneity. 
Specifically, CaBaFL maintains a hierarchical cache data structure on the server that can: 
i) mitigate the straggler problem caused by device heterogeneity using our proposed hierarchical cache-based aggregation mechanism, and ii) achieve stable  training  convergence and high global model accuracy
by properly placing models in different hierarchies of the cache.
Moreover, by using our feature balance-guided device selection strategy, CaBaFL can alleviate the performance deterioration caused by data imbalance. 
Comprehensive experimental results demonstrate that CaBaFL can achieve much better inference performance compared with state-of-the-art heterogeneous FL methods within both IID and non-IID scenarios.

%

\section{Acknowledgement}
This work was supported by Natural Science Foundation of China (62272170),   and ``Digital Silk Road'' Shanghai International Joint Lab of Trustworthy Intelligent Software (22510750100), Shanghai Trusted Industry Internet Software Collaborative Innovation Center, the National Research Foundation, Singapore, and the Cyber Security Agency under its National Cybersecurity R\&D Programme (NCRP25-P04-TAICeN).
Any opinions, findings and conclusions or recommendations expressed in this material are those of the author(s) and do not reflect the views of National Research Foundation, Singapore and Cyber Security Agency of Singapore.


\bibliographystyle{IEEEtran}
\bibliography{sample-base}


\end{document}